\documentclass[10pt,twocolumn,letterpaper]{article}
\usepackage{cvpr}              

\usepackage{arydshln}
\usepackage{dashrule}

\usepackage{dsfont}
\usepackage{dashrule}

\usepackage{multirow}
\usepackage{colortbl}
\definecolor{cvprblue}{rgb}{0.21,0.49,0.74}
\usepackage[pagebackref,breaklinks,colorlinks,allcolors=cvprblue]{hyperref}

\def\paperID{928} 
\def\confName{CVPR}
\def\confYear{2025}

\title{Unlocking Generalization Power in LiDAR Point Cloud Registration}
\author{Zhenxuan Zeng$^{1,2}$ , Qiao Wu$^{1,2}$, Xiyu Zhang$^{1,2}$, 
Lin Yuanbo Wu$^{3}$, Pei An$^{4}$ \\
Jiaqi Yang$^{1,2}$\thanks{Corresponding author}, Ji Wang$^{1,2\ast}$, Peng Wang$^{1,2}$
\\
$^1$School of Computer Science, Northwestern Polytechnical University, China.\\
$^2$Ningbo Institute, Northwestern Polytechnical University, China.\\
$^3$Department of Computer Science, Swansea University, United Kingdom.\\
$^4$HuaZhong University of Science and Technology, China.\\
{\tt\small \{zengzhenxuan,qiaowu\}@mail.nwpu.edu.cn
\tt\small \{jqyang, j.wang, peng.wang\}@nwpu.edu.cn
}}

\begin{document}
\maketitle

\begin{abstract}
\label{sec:abstract}
\indent
In real-world environments, a LiDAR point cloud registration method with robust generalization capabilities (across varying distances and datasets) is crucial for ensuring safety in autonomous driving and other LiDAR-based applications. However, current methods fall short in achieving this level of generalization. To address these limitations, we propose UGP, a pruned framework designed to enhance generalization power for LiDAR point cloud registration. The core insight in UGP is the elimination of cross-attention mechanisms to improve generalization, allowing the network to concentrate on intra-frame feature extraction. Additionally, we introduce a progressive self-attention module to reduce ambiguity in large-scale scenes and integrate Bird’s Eye View (BEV) features to incorporate semantic information about scene elements. Together, these enhancements significantly boost the network’s generalization performance. We validated our approach through various generalization experiments in multiple outdoor scenes. In cross-distance generalization experiments on KITTI and nuScenes, UGP achieved state-of-the-art mean Registration Recall rates of 94.5\% and 91.4\%, respectively. In cross-dataset generalization from nuScenes to KITTI, UGP achieved a state-of-the-art mean Registration Recall of 90.9\%. Code will be available at \textcolor{cvprblue}{\href{https://github.com/peakpang/UGP}{https://github.com/peakpang/UGP}}
\end{abstract}
    
\section{Introduction}
\label{sec:intro}

\begin{figure}[t]
    \centering
    \includegraphics [width=1\linewidth]{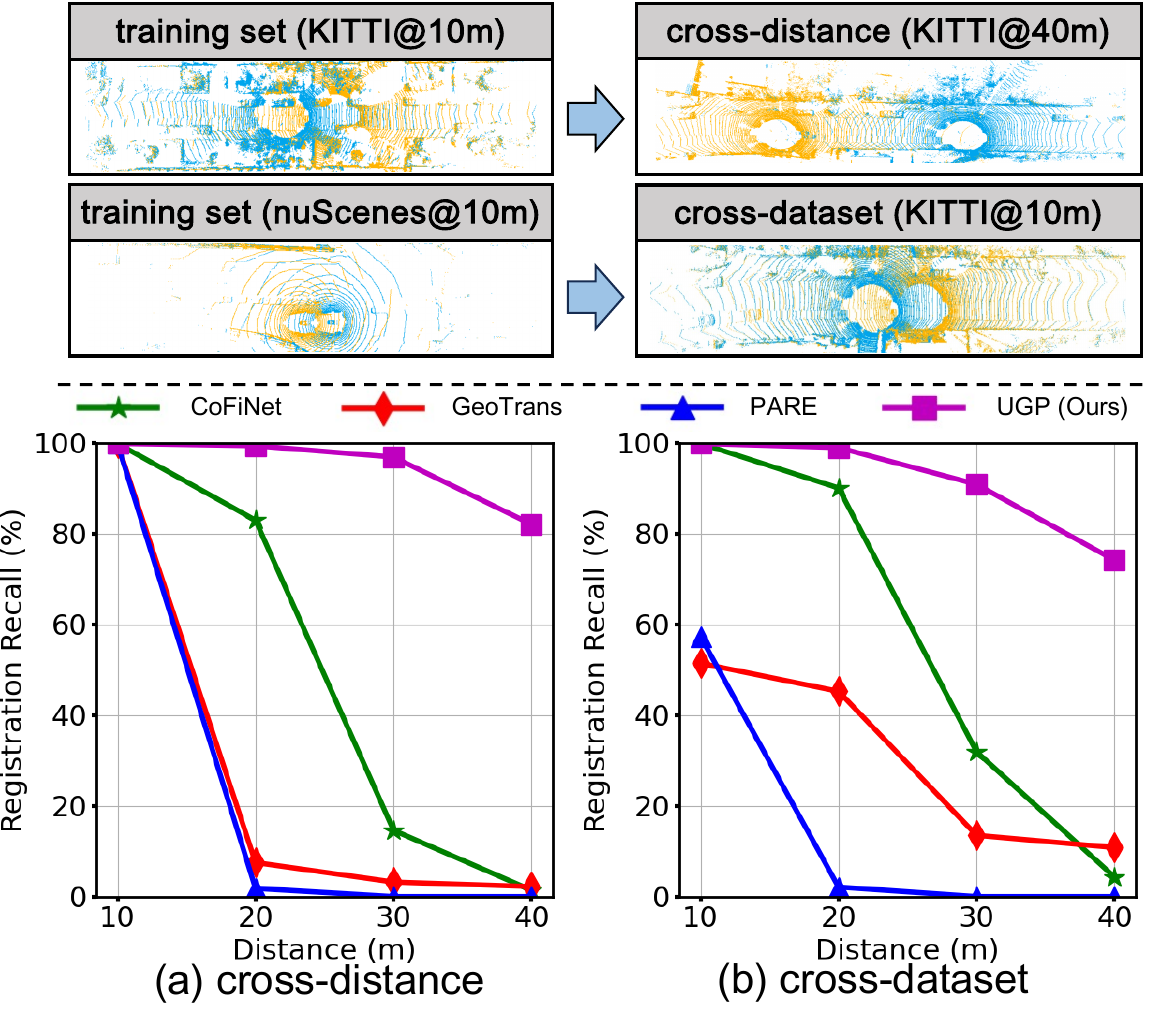}
    \caption{Illustration of the generalization performance of leading methods \cite{yu2021cofinet, qin2022geometric, yao2024pare} in cross-distance and cross-dataset. (a) Cross-distance generalization: train on KITTI@10m, test on KITTI@40m. (b) Cross-dataset generalization: train on nuScenes@10m \cite{nuscenes}, test on KITTI@10m. These methods experience substantial performance degradation due to poor generalization. In contrast, our method achieves the best registration results on both cross-distance and cross-dataset.}
    \label{fig1}
\end{figure}

LiDAR point cloud registration is essential in applications such as autonomous driving \cite{guo20143d} and SLAM \cite{mur2015orb,mur2017orb,shan2020lio}. The dynamic nature of LiDAR scenes often necessitates pairwise registration of point clouds that exhibit 1) cross-distance variations (\eg, point clouds captured at different speeds or times) and 2) cross-dataset variations (\eg, point clouds collected in distinct environments). To tackle these real-world challenges, this research seeks to develop a LiDAR point cloud registration framework that can effectively generalize across both cross-distance and cross-dataset variations.

Recently, learning-based methods for point cloud registration have gained prominence \cite{deng2018ppfnet,huang2021predator,ao2021spinnet,bai2020d3feat,cao2021pcam,GeDi2022,gojcic2019perfect,yu2023rotation}. However, most of these methods focus on enhancing performance under same-distance or same-dataset conditions, with only a few addressing cross-dataset generalization. Notably, SpinNet \cite{ao2021spinnet} and BUFFER \cite{ao2023buffer} use patch-wise feature extraction, which provides robustness against noise and occlusion. Nonetheless, to our knowledge, none of existing methods has systematically investigated generalization performance for LiDAR point cloud registration. Unlike evenly distributed indoor RGB-D point clouds, LiDAR point clouds exhibit uneven distribution where the point density decreases significantly as the distance from the sensor increases. This introduces unique challenges: 1) cross-distance variations alter the overlap rate between point cloud pairs and cause density changes within the overlap region, and 2) cross-dataset variations lead to differences in data characteristics (\eg, 64-line vs. 32-line LiDAR) and feature distributions (\eg, environmental shifts). We observe that state-of-the-art (SOTA) methods suffer noticeable performance degradation under both cross-distance and cross-dataset conditions. As shown in Fig.~\ref{fig1}, CoFiNet \cite{yu2021cofinet}, GeoTrans \cite{qin2022geometric}, and PARE \cite{yao2024pare} are coarse-to-fine methods that use the transformer to improve registration performance under the same distance or same dataset conditions. Among these approaches, cross-attention is extensively used to model the geometric consistency across the two point clouds. However, the effectiveness of this module relies on an implicit assumption: \textit{consistency in the representation of the same structure across the two point clouds.} In dynamic scenarios where cross-distance and cross-dataset generalization challenges arise, this implicit assumption often does not hold due to the inconsistent density distribution of LiDAR point clouds.

To tackle these challenges, we propose a LiDAR point cloud registration framework with strong generalization power (\textbf{UGP}). The framework’s key insight is eliminating the cross-attention module to \textbf{U}nlock \textbf{G}eneralization \textbf{P}ower and enable the network to focus on intra-frame feature extraction. First, to reduce the impact of inconsistent geometric representations in inter-frame, we eliminate the cross-attention module. This encourages the network to concentrate on intra-frame spatial information. Second, to reduce feature ambiguity in LiDAR scenes and capture finer local feature associations, we introduce a progressive self-attention module. This module gradually increases the attention range of superpoints, enabling them to prioritize local spatial information and construct a multi-scale spatial representation. Lastly, recognizing the importance of scene-element-level (\eg, roads, corners) semantic information in reducing scene ambiguity, we incorporate Bird Eye View (BEV) features to improve accuracy during the coarse matching stage. In summary, our main contributions are as follows:

\begin{itemize}
        \item We reveal that inconsistent geometric representations in LiDAR scenarios cause the cross-attention module to limit network generalization capability. An effective way to unlock this is to eliminate the cross-attention module.

        \item We also propose a progressive self-attention module and a BEV feature extraction module to more effectively focus on fine local space associations and capture semantically rich information, reducing ambiguity in point clouds and enhancing generalization.
        
        \item In cross-distance generalization from 10m to 40m, UGP achieved 82.0\% (+20.8\%) registration recall (RR) on KITTI and 72.3\% (+57.1\%) RR on nuScenes, outperforming BUFFER \cite{ao2023buffer}. In cross-dataset generalization from nuScenes to KITTI, our method achieved a state-of-the-art mean RR of 90.9\% (+6.2\%) over BUFFER \cite{ao2023buffer}.
        
\end{itemize}

\section{Related Work}
\label{sec:Related Work}
\begin{figure*}[t]
    \centering
    \includegraphics [width=1\linewidth]{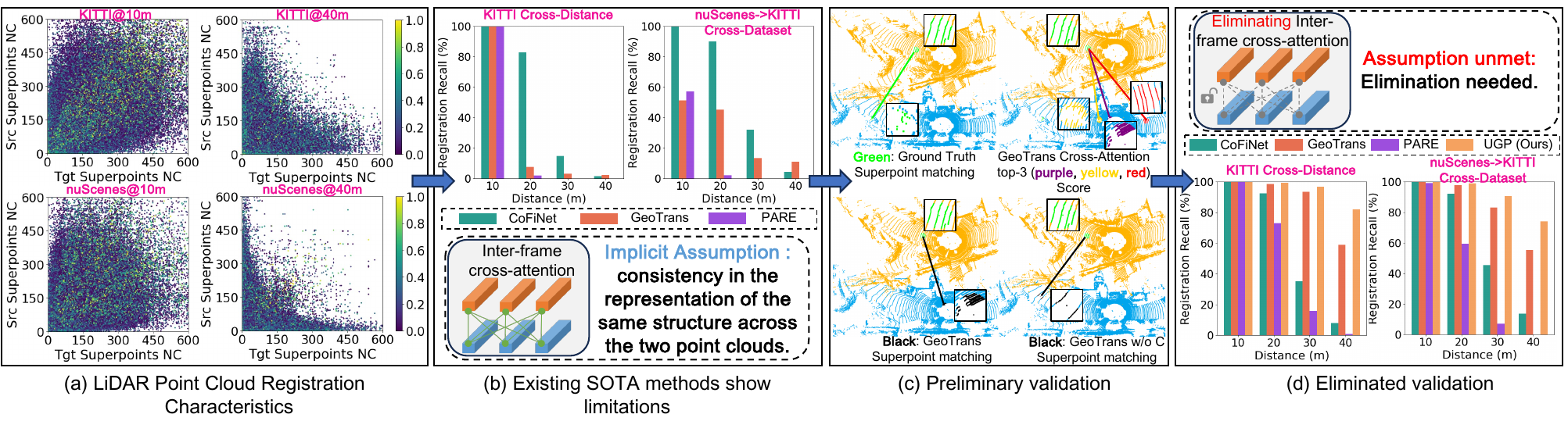}
    \caption{\textbf{Motivation.} We conducted a statistical analysis of the matching characteristics in LiDAR scenes and identified the limitations of existing methods. This was followed by preliminary visualizations and elimination experiments to validate our findings. (a) Each point in the figure represents a ground truth corresponding superpoint pair. The position of each point indicates the neighborhood count (NC) of the superpoint within a radius of $r=2.4m$ in both the source (src) and target (tgt) point clouds, and the color represents the overlap degree of the corresponding superpoint pairs after rotation by the ground truth transformation. (b) Performance of existing methods in cross-distance (upper left) and cross-dataset (upper right). (c) GeoTrans \cite{qin2022geometric} cannot match superpoints (cross-distance, from KITTI@10m to KITTI@20m). (d) The performance of existing methods after eliminating cross-attention, and UGP (ours).
    }
    \label{fig:motivation}
\end{figure*}
\noindent\textbf{Learning-based Point Cloud Registration.} Learning-based point cloud registration methods have become a powerful approach for aligning point clouds. Current methods often employ a coarse-to-fine strategy \cite{yu2021cofinet,qin2022geometric,yu2023rotation,yu2023peal,chen2024dynamic,yao2024pare,mu2024colorpcr,xiong2024speal}, which loosens the constraints of strict point-to-point matching in favor of patch-level matching. This approach reduces repeatability requirements, significantly enhancing registration accuracy. However, these methods rely heavily on the data distribution of training point cloud pairs, which limits their generalization capability. Following this line, point-wise methods \cite{choy2019fully,bai2020d3feat} utilize architectures such as Minkowski \cite{Minkowski2019} and KPConv \cite{thomas2019kpconv} to extract dense point features, while patch-wise methods \cite{deng2018ppfnet,gojcic2019perfect,ao2021spinnet,GeDi2022,ao2023buffer} capture features from the local regions around key points through carefully designed networks. The robustness of local patches to noise and occlusion gives these patch-based approaches a promising generalization potential. Recent studies \cite{liu2023apr,liu2023density, liu2024extend} have focused on point cloud registration across varying distances. However, these approaches rely solely on visible full-range data and require carefully designed multi-frame training to ensure registration performance within the visible range.

\noindent\textbf{Point Cloud Registration with Transformers.} The Transformer architecture \cite{vaswani2017attention} introduced an attention mechanism that captures global dependencies and has been widely used for point cloud registration. For example, CoFiNet \cite{yu2021cofinet} uses vanilla attention mechanism to establish global geometric consistency, while GeoTrans \cite{qin2022geometric} integrates geometric structure to enhance feature robustness. PEAL \cite{yu2023peal} employs one-way attention with overlap priors to reduce ambiguity from non-overlapping points. DCATr \cite{chen2024dynamic} limits attention to specific cues to minimize irrelevant interference. Unlike these methods, we propose a progressive self-attention mechanism that gradually expands the spatial attention range, prioritizing finer local spatial structure associations, reducing ambiguity in large scenes, and improving registration accuracy.

\noindent\textbf{Multi-modal Point Cloud Registration.} Incorporating multi-modal information for point cloud understanding has gained significant traction in recent research. For example, RPVNet \cite{xu2021rpvnet} enhances segmentation accuracy by integrating the range image modality. Methods such as BYOC \cite{el2021bootstrap}, PCRCG \cite{zhang2022pcr}, and IGReg \cite{xu2024igreg} fuse image features with point clouds to support registration tasks. Similarly, PEAL2D \cite{yu2023peal} utilizes a 2D image to derive an overlap prior, aiding in point cloud registration. BEV images, widely adopted in 3D object detection \cite{Yang_2018_CVPR} and place recognition~\cite{luo2023bevplace}, are particularly effective due to their clear global edge features. To enhance matching accuracy in the coarse stage, we explicitly incorporate BEV images into point cloud registration.

\section{Motivation}
\label{sec:Motivation}
\begin{figure*}[t]
    \centering
    \includegraphics [width=1\linewidth]{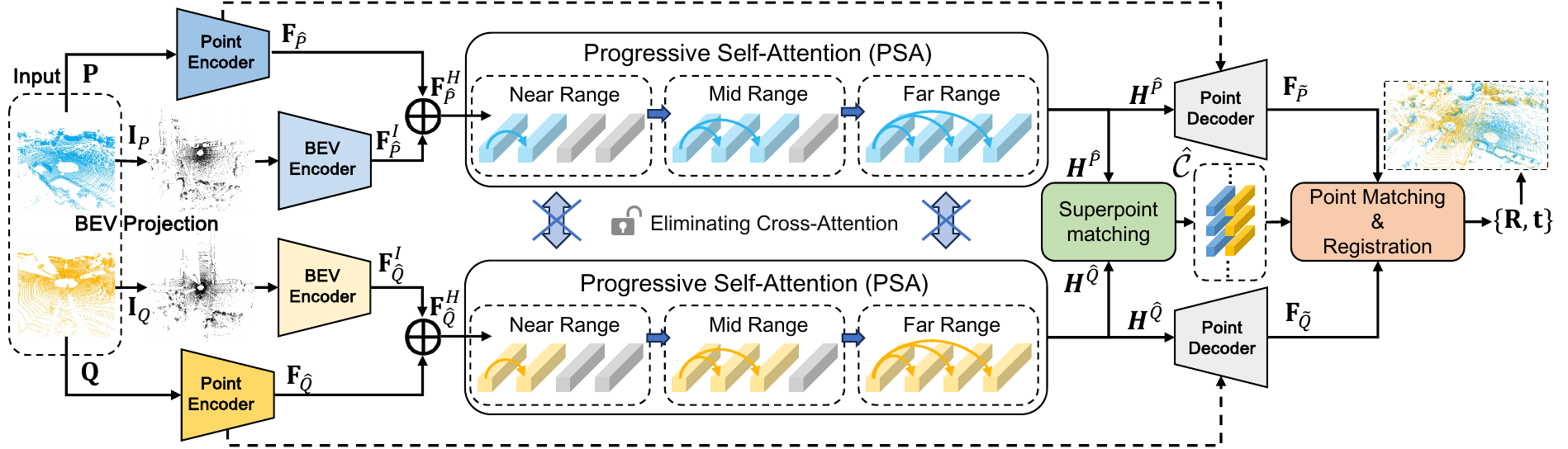}
    \caption{\textbf{Overview of UGP.} The input point cloud is first projected to obtain the corresponding BEV image. Then, the point cloud and BEV image are fed into the Point Encoder and BEV encoder, respectively, for downsampling and feature extraction. In the superpoint matching stage, the indexing relationship between superpoints and BEV is used to fuse point-level features and image features. The fused features are used as input to the progressive self-attention module to extract robust and consistent intra-frame features. Finally, the superpoint matching results are propagated to dense points for dense matching, enabling the recovery of the rigid transformation.
}
    \label{fig:UGP}
\end{figure*}
In real-world dynamic scenarios, we face the challenge of registering point clouds with 1) cross-distance variations (\eg, point clouds captured at different speeds or times) and 2) cross-dataset variations (\eg, point clouds collected in distinct environments or using different LiDAR types). This challenge requires the LiDAR point cloud registration framework to possess cross-distance and cross-dataset generalization power. To address this, we analyze the matching characteristics under both challenges and find that existing methods have limited performance. Finally, we propose targeted improvements that effectively address these existing challenges.

\noindent\textbf{a) Inherent Challenges with LiDAR Point Cloud Registration.} Fig. \ref{fig:motivation} (a) highlights the challenges in LiDAR point cloud registration by visualizing data distributions of ground truth matching point pairs across varying distances and datasets. 1) At closer distances (\eg, KITTI@10m or nuScenes@10m), there is a high number of matched pairs with consistent density (clustered near $y=x$) and a high overlap score. As the distance increases (\eg, KITTI@40m vs. KITTI@10m or nuScenes@40m vs. nuScenes@10m), the proportion of pairs with inconsistent density grows. Additionally, increased distance brings changes in perspective and occlusion, reducing overlap rates even among points with similar density. 2) Cross-dataset comparisons reveal differences in data characteristics, such as the shift from nuScenes (32-line LiDAR) to KITTI (64-line LiDAR).

\noindent\textbf{b) Implicit Assumption of Cross-Attention Based Methods.} The effectiveness of SOTA methods such as CoFiNet \cite{yu2021cofinet}, GeoTrans \cite{qin2022geometric}, and PARE \cite{yao2024pare} relies on an implicit assumption: \textit{consistency in the representation of the same structure across two point clouds}. However, cross-attention cannot adapt to variations in the consistency representation of the same structure across different distances and datasets, limiting the network’s generalization performance, as shown in Fig. \ref{fig:motivation} (b).

\noindent\textbf{c) Cross-Attention Elimination.} To demonstrate the limitations of cross-attention, we present results from GeoTrans~\cite{qin2022geometric} trained on KITTI@10m and tested on KITTI@20m, as shown in Fig.~\ref{fig:motivation} (c). First, we show a challenging superpoint matching example on the ground, where uneven LiDAR distribution and limited geometric detail make accurate matching difficult. As the distance between point cloud pairs increases (\eg, green ground-truth matches at 20m), variations in density and occlusion worsen matching accuracy. Cross-attention-based methods like GeoTrans~\cite{qin2022geometric}, fail in cross-distance matching by relying on density-similar but incorrect matches. Visualization of GeoTrans~\cite{qin2022geometric}’s top-3 score areas from its final cross-attention layer shows that it primarily captures similarly dense structures and struggles to adapt to shifts in data distribution, resulting in outlier matches. By eliminating cross-attention, superpoint matching refocuses on correct areas near the ground truth. Preliminary validation followed by elimination experiments (Fig. \ref{fig:motivation} (d)) confirms that eliminating cross-attention greatly enhances model generalization.

\noindent\textbf{d) Boosting the Robustness and Consistency of Intra-Frame Feature Learning.} After eliminating the cross-attention module, we encourage the network to concentrate on intra-frame spatial information. A vanilla self-attention~\cite{vaswani2017attention} forces each point to interact with all other points equally. However, this leads to feature ambiguity due to irrelevant distant points. To improve the robustness and consistency of intra-frame feature extraction, we propose a progressive self-attention. In the initial layer, the attention range is restricted to the local space, preserving fine-grained local information. In subsequent layers, the attention range is gradually expanded to incorporate the global context. This progressive approach reduces feature ambiguity and enhances the model's ability to effectively capture both local and global structure. Besides, point-based backbones such as KPConv~\cite{thomas2019kpconv} are difficult to establish a association between local geometric structures and global background information \cite{liu2019lpd, xu2021rpvnet}. This further worsens the ambiguity of the features. To this end, we introduce BEV which can provide additional semantic information of scene elements (\eg, roads, corners) to the local geometric features extracted by KPConv.

\section{Method}
\label{sec:method}

Given two sets of partially overlapped point clouds $\mathbf{P}\in\mathbb{R}^{N\times3}$ and $\mathbf{Q}\in\mathbb{R}^{M\times3}$, our objective is to recover the optimal rigid transformation between them, denoted as $\mathbf{T}=\left\{\mathbf R \in SO(3),\mathbf t\in \mathbb{R}^{3}\right\}$, where $\mathbf{R}$ represents rotation and $\mathbf{t}$ represents translation. As illustrated in Fig. \ref{fig:UGP}, we adopt a coarse-to-fine strategy and eliminate the cross-attention module. First, the point cloud is directly projected to obtain the BEV image (Sec. \ref{sec:BEV Projection}). Then, the Point-Encoder and BEV-Encoder are used to extract the scale features of the point cloud and BEV image. Next, the patch features of the point cloud and image are fused to obtain the superpoint features (Sec. \ref{sec:Superpoint Feature Extraction and Fusion}). The superpoint matching module eliminates the cross-attention module and utilizes only the progressive self-attention to extract superpoint features, thereby obtaining superpoint correspondences (Sec. \ref{sec:Superpoint Matching Module}). Finally, we follow GeoTransformer \cite{qin2022geometric}, using the same point matching module to refine the superpoint matching and recover the alignment transformations with the LGR \cite{qin2022geometric} approach (Appendix Sec.~\ref{sec:Method Details}).

\subsection{BEV Projection}  \label{sec:BEV Projection}
In LiDAR point cloud registration, the significant global edge and texture feature are important for superpoint matching. From a 2D perspective, the BEV representation of the point cloud provides a global view of the environment and offers important semantic information about scene elements. These prominent features reduce the ambiguity of the scene and improve feature consistency.

Given a point cloud $\mathbf{P}$ $ = \left \{(x_i, y_i, z_i) \mid i = 1, 2, \dots, N \right \}$, where $(x_i, y_i, z_i)$ represents the coordinates of each point, we project into a BEV image $\mathbf I \in \mathbb{R}^{H \times W}$, where \( H \) and \( W \) denote the height and width of the BEV image, respectively. Assuming that the ground area covered by this aerial view is $[x_{\min}, x_{\max}]$ and $[y_{\min}, y_{\max}]$, the projection process can be expressed by the following formula :
\begin{equation}\small
u_i = \left\lfloor \frac{x_i - x_{\min}}{x_{\max} - x_{\min}} \cdot H \right\rfloor,
v_i = \left\lfloor \frac{y_i - y_{\min}}{y_{\max} - y_{\min}} \cdot W \right\rfloor,
\end{equation}
where $(u_i, v_i)$ is the pixel coordinate of point $(x_i, y_i, z_i)$ in the BEV image. To capture the global structure, we fill each pixel position with a gray value of $\mathbf I (u_i, v_i) = 1$.

\subsection{Superpoint Feature Extraction and Fusion} \label{sec:Superpoint Feature Extraction and Fusion}
Our feature extraction module consists of a Point-Encoder and a BEV-Encoder. The Point-Encoder employs a multi-level KPConv structure \cite{thomas2019kpconv}, with voxel downsampling applied at each layer to decrease the number of points and ultimately extract superpoint features. The BEV-Encoder uses a multi-level ResNet structure \cite{resnet}, applying 2D max pooling at each layer to downsample the image and extract patch-level features. Finally, we fuse the two features to obtain richer superpoints feature expressions.

Specifically, given the input point clouds $\mathbf P$ and $\mathbf Q$, along with their corresponding BEV images $\mathbf I_P$ and $\mathbf I_Q$, the Point-Encoder outputs the downsampled superpoints  $\mathbf{\hat{P}}$ and $\mathbf{\hat{Q}}$, as well as their corresponding features  $\mathbf{F}_{\hat{P}} \in \mathbb{R}^{\hat{n}   \times \hat{d}}$ and $\mathbf{F}_{\hat{Q}} \in \mathbb{R}^{\hat{m} \times \hat{d}}$, where \(\hat{d}\) denotes the feature dimension of each superpoint. At the same time, the BEV-Encoder generates the patch features $\mathbf F_{\hat{P}}^{I} \in \mathbb{R}^{H’ \times W’ \times d’}$ and $\mathbf F_{\hat{Q}}^{I} \in \mathbb{R}^{H’ \times W’ \times d’}$, where \(H'\) and \(W'\) represent the height and width of the downsampled BEV image, and \(d'\) denotes the feature dimension for each patch in the BEV image. Afterwards, feature fusion is performed based on the indexing relationship between superpoints and BEV patches, where the features are concatenated to combine 3D geometry and 2D texture information. This results in the fused features $\mathbf F_{\hat{P}}^{H}$  and  $\mathbf F_{\hat{Q}}^{H}$ for the superpoints $\mathbf{\hat{P}}$ and $\mathbf{\hat{Q}}$, respectively.

\subsection{Superpoint Matching Module} \label{sec:Superpoint Matching Module}
For the coarse-to-fine method, superpoint matching accuracy directly determines final registration performance.

\noindent\textbf{Eliminating Cross-Attention to Concentrate on Intra-Frame Feature Learning.}
The distributional characteristics of LiDAR data differ from indoor RGB-D point clouds, causing inconsistencies in geometric representations across distances and datasets and limiting network generalization. To improve the framework’s generalization performance, we adopt a simple, effective approach based on the observations in Sec. \ref{sec:Motivation}: Eliminating the cross-attention module and retaining self-attention to focus on intra-frame feature extraction, enhancing generalization.

In the following, we describe the process for computing the feature for superpoint $\mathbf{\hat{P}}$ with self-attention, which similarly applies to $\mathbf{\hat{Q}}$. Given the input feature matrix  $\mathbf{X} \in \mathbb{R}^{\hat{n} \times d_{t}}$,  where \(d_{t}\) represents the dimension of each input feature, the output feature matrix  $\mathbf{Z} \in \mathbb{R}^{\hat{n} \times d_{t}}$  is obtained as the weighted sum of the projected input features:
\begin{equation}
\mathbf z_{i}=\sum_{j=1}^{\hat{n}} a_{i,j} \left (\mathbf{x}_{j} \mathbf{W}^{V}\right),
\end{equation}
where the attention weights $a_{i,j}$ are obtained by applying a row-wise softmax to the attention scores $e_{i,j}$, and the attention scores $e_{i,j}$ computed as:
\begin{equation}
e_{i, j}=\frac{\left(\mathbf{x}_{i} \mathbf{W}^{Q}\right)\left(\mathbf{x}_{j} \mathbf{W}^{K}+\mathbf{r}_{i, j} \mathbf{W}^{R}\right)^{T}}{\sqrt{d_{t}}}.
\end{equation}
Here, $\mathbf{r}_{i, j}$ refers to the geometric structure embedding proposed in GeoTransformer \cite{qin2022geometric}. The terms $\mathbf{W}^{Q}$,$\mathbf{W}^{K}$,$\mathbf{W}^{V}$,$\mathbf{W}^{R}\in \mathbb{R}^{d_{t}\times d_{t}} $ are the respective projection matrices for queries, keys, values and geometric structure embeddings. 

\begin{figure}[t]
    \centering
    \includegraphics [width=1\columnwidth]{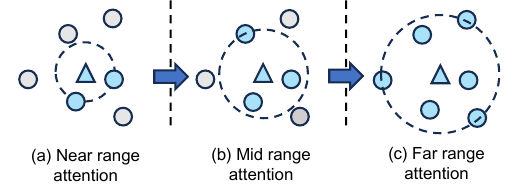}
    \caption{Illustration of our proposed progressive self-attention. In the initial layer, the triangular points prioritize attention to the local space around them, and the attention range is gradually expanded in subsequent layers}
    \label{fig:range}
\end{figure} 
\begin{table*}[t] \centering
    \resizebox{\linewidth}{!}{
    \begin{tabular}{l|ccccc|ccccc|ccccc|ccccc|l}
        \toprule
        \textbf{Model} & \multicolumn{5}{c}{\textbf{Train on KITTI@10m}} & \multicolumn{5}{c}{\textbf{KITTI@20m}} & \multicolumn{5}{c}{\textbf{KITTI@30m}} & \multicolumn{5}{c}{\textbf{KITTI@40m}} &  \\
        \cmidrule(lr){2-6} \cmidrule(lr){7-11} \cmidrule(lr){12-16} \cmidrule(lr){17-21}
        & RRE & RTE & RRE$^{*}$ & RTE$^{*}$ & RR & RRE & RTE & RRE$^{*}$ & RTE$^{*}$ & RR & RRE & RTE & RRE$^{*}$ & RTE$^{*}$ & RR & RRE & RTE & RRE$^{*}$ & RTE$^{*}$ & RR & mRR \\
        \midrule
        FCGF \cite{choy2019fully} & \textbf{0.213} & \underline{0.064} & 0.864 & 0.185 & \underline{98.9} & \textbf{0.358} & \textbf{0.121} & 5.812 & 1.477 & 92.1 & 1.051 & 0.339 & 11.706 & 10.514 & 64.1 & 1.477 & 0.767 & 19.855 & 30.598 & 18.7 & 68.5\\
        SpinNet \cite{ao2021spinnet} & 0.515 & 0.104 & 0.948 & 0.322 & 97.7 & 1.436 & 0.237 & 9.418 & 7.833 & 58.0 & 2.018 & 0.638 & 22.107 & 28.821 & 1.1 & -- & -- & 23.191 & 39.544 & 0.0 & 39.2\\
        Predator \cite{huang2021predator} & 0.282 & 0.070 & 0.282 & 0.076 & \textbf{99.8} & 1.602 & 0.324 & 16.438 & 12.403 & 30.2 & 1.903 & 0.419 & 28.594 & 28.054 & 1.1 & -- & -- & 30.605 & 38.637 & 0.0 & 32.8 \\
        CoFiNet \cite{yu2021cofinet}& 0.344 & 0.083 & 0.357 & 0.089 & \textbf{99.8} & 1.133 & 0.256 & 3.758 & 2.924 & 82.9 & 1.862 & 0.557 & 18.521 & 24.992 & 14.6 & 2.054 & 0.807 & 25.716 & 39.071 & 1.4 & 49.7\\
        GeoTrans \cite{qin2022geometric} & 0.244 & 0.069 & 0.301 & 0.075 & \textbf{99.8} & 1.640 & 0.395 & 26.555 & 24.533 & 7.5 & 1.355 & 0.503 & 45.092 & 34.987 & 3.2 & 1.431 & 0.618 & 46.972 & 45.065 & 2.2 & 28.2 \\
        BUFFER \cite{ao2023buffer}& 0.259 & 0.072 & \underline{0.262} & 0.079 & \textbf{99.8} & 0.437 & \underline{0.131} & \underline{0.475} & 0.345 & \underline{98.6} & 0.719 & \textbf{0.230} & \underline{2.290} & 1.910 & \underline{93.5} & 1.050 & \textbf{0.364} & \underline{14.221} & 15.811 & \underline{61.2} & \underline{88.3} \\
        PARE \cite{yao2024pare}& 0.239 & \textbf{0.054} & \textbf{0.242} & \textbf{0.061} & \textbf{99.8} & 1.284 & 0.401 & 19.192 & 27.340 & 1.8 & -- & -- & 23.682 & 37.837 & 0.0 & -- & -- & 29.139 & 47.570 & 0.0 & 25.4\\
        \cdashline{0-21}
        CoFiNet w/o C & 0.366 & 0.085 & 0.379 & 0.109 & \textbf{99.8} & 0.813 & 0.182 & 2.305 & 1.504 & 92.5 & 1.734 & 0.490 & 14.172 & 18.061 & 35.1 & 1.970 & 0.785 & 27.077 & 36.130 & 7.9 & 58.8\\
        GeoTrans w/o C & \underline{0.238} & 0.066 & 0.295 & \underline{0.073} & \textbf{99.8} & 0.426 & 0.146 & 0.501 & \underline{0.225} & \underline{98.6} & \underline{0.699} & \underline{0.285} & 4.808 & \underline{1.889} & \underline{93.5} & \textbf{0.949} & \underline{0.476} & 32.318 & \underline{14.096} & 59.0 & 87.7\\
        PARE w/o C & 0.289 & 0.071 & 0.292 & 0.078 & \textbf{99.8} & 0.924 & 0.288 & 9.548 & 4.627 & 73.0 & 1.330 & 0.574 & 38.783 & 23.824 & 15.7 & 3.660 & 1.456 & 47.526 & 39.225 & 0.7 & 47.3\\
        \cdashline{0-21}
        UGP(Ours) & 0.245 & 0.071 & 0.298 & 0.078 & \textbf{99.8} & \underline{0.393} & 0.147 & \textbf{0.449} & \textbf{0.221} & \textbf{99.3} & \textbf{0.677} & 0.290 & \textbf{2.127} & \textbf{0.805} & \textbf{96.8} & \underline{0.975} & 0.494 & \textbf{13.095} & \textbf{6.168} & \textbf{82.0} & \textbf{94.5}\\
        \midrule
        \textbf{Model} & \multicolumn{5}{c}{\textbf{Train on nuScenes@10m}} & \multicolumn{5}{c}{\textbf{nuScenes@20m}} & \multicolumn{5}{c}{\textbf{nuScenes@30m}} & \multicolumn{5}{c}{\textbf{nuScenes@40m}} & \\
        \cmidrule(lr){2-6} \cmidrule(lr){7-11} \cmidrule(lr){12-16} \cmidrule(lr){17-21}
        & RRE & RTE & RRE$^{*}$ & RTE$^{*}$ & RR & RRE & RTE & RRE$^{*}$ & RTE$^{*}$ & RR & RRE & RTE & RRE$^{*}$ & RTE$^{*}$ & RR & RRE & RTE & RRE$^{*}$ & RTE$^{*}$ & RR & mRR \\
        \midrule
        FCGF \cite{choy2019fully} & 0.338 & \textbf{0.176} & 0.396 & 0.374 & 97.4 & \underline{0.511} & 0.343 & 2.524 & 3.105 & 84.4 & 1.015 & 0.492 & \underline{6.801} & \underline{12.178} & \underline{53.5} & 1.560 & 0.519 & \underline{12.496} & 28.355 & \underline{19.7} & \underline{63.8} \\
        Predator \cite{huang2021predator} & 0.365 & \underline{0.178} & 0.365 & \textbf{0.178} & \textbf{100} & 1.640 & \textbf{0.312} & 22.617 & 10.038 & 15.8 & 1.855 & \textbf{0.279} & 25.102 & 20.76 & 5.3 & 1.809 & \underline{0.175} & 31.424 & 29.966 & 5.1 & 31.6\\
        CoFiNet \cite{yu2021cofinet}& 0.461 & 0.195 & 0.625 & 0.416 & 97.0 & 1.370 & 0.482 & 18.662 & 7.398 & 58.0 & 1.810 & 0.556 & 33.388 & 23.109 & 12.7 & 1.923 & 0.365 & 32.966 & 33.459 & 6.5 & 43.6\\
        GeoTrans \cite{qin2022geometric}& \textbf{0.259} & \underline{0.178} & \textbf{0.259} & \textbf{0.178} & \textbf{100} & 0.733 & 0.412 & 9.099 & 5.059 & 57.1 & \underline{0.734} & 0.382 & 22.755 & 19.452 & 9.6 & \underline{0.571} & 0.294 & 29.104 & 29.317 & 7.2 & 43.5\\
        BUFFER \cite{ao2023buffer}& 0.340 & \underline{0.178} & 0.390 & 0.394 & 97.5 & 0.528 & 0.347 & \underline{0.866} & \underline{2.917} & \underline{84.7} & 0.918 & 0.523 & 9.901 & 14.183 & 46.6 & 1.073 & 0.567 & 25.865 & 30.811 & 15.2 & 61.0 \\
        PARE \cite{yao2024pare}& 0.449 & 0.190 & 0.475 & 0.202 & \underline{99.7} & 0.906 & 0.445 & 6.834 & 9.638 & 9.2 & \textbf{0.674} & \underline{0.365} & 10.530 & 19.084 & 5.3 & \textbf{0.346} & \textbf{0.163} & 13.746 & \underline{28.349} & 4.6 & 29.7\\
        UGP(Ours) & \underline{0.270} & 0.179 & \underline{0.270} & \underline{0.179} & \textbf{100} & \textbf{0.467} & \underline{0.338} & \textbf{0.646} & \textbf{0.415} & \textbf{99.4} & 0.780 & 0.460 & \textbf{1.284} & \textbf{1.456} & \textbf{93.9} & 1.142 & 0.636 & \textbf{10.532} & \textbf{8.641} & \textbf{72.3} & \textbf{91.4}\\
        \bottomrule
    \end{tabular}
}
    \caption{\textbf{Cross-distance generalization experiments.}  For both the KITTI and nuScenes datasets, \textbf{we train at 10m and then test at 10m and farther distances at 20m, 30m, and 40m.} RRE and RTE denote the error for successfully matched point cloud pairs, while RRE$^*$ and RTE$^*$ reflect the error for all point cloud pairs, providing a more comprehensive evaluation. The final column shows the mean Registration Recall.}
    \label{tab:cross-distance}
\end{table*}
\noindent\textbf{Progressive Self-Attention Module.}
In real-world scenarios, local contextual information plays a crucial role in point cloud feature extraction. Existing methods \cite{yu2021cofinet,qin2022geometric,yu2023peal,chen2024dynamic,yao2024pare} apply global self-attention directly to the input point cloud features. Although self-attention can capture long-range global dependencies, this approach is more susceptible to noise in large-scale scenes, which aggravates the ambiguity in feature matching. To reduce the matching ambiguity in LiDAR scenes, we introduce a progressive self-attention module, as shown in Fig.~\ref{fig:range}. This module dynamically adjusts the attention range in a progressive manner, based on the distances between each point and other points in the global context. In the initial layer ($L=1$), the module targets local association refinement by focusing on nearby superpoints, effectively reducing the ambiguity that can arise from initial global attention. In subsequent layers ($L > 1$), the attention range is gradually expanded, enabling the model to build multi-scale spatial representations of superpoints. This stepwise approach enhances the robustness of superpoint features.

Next, we will detail the computation process for $\mathbf{\hat{P}}$ at layer $L = k$, which similarly applies to $\mathbf{\hat{Q}}$. To implement progressive, the masked attention scores $e_{i,j}^{(L=k)}$ are computed as:
\begin{equation}
e_{i, j}^{(L=k)} = \mathbf M_{i, j}^{(L=k)} \cdot e_{i, j},
\end{equation}
where $\mathbf{M}_{i, j}^{(L=k)}$ is a layer-specific mask designed to filter attention scores based on the Euclidean distance between points. The mask $\mathbf{M}_{i, j}^{(L=k)}$ is computed by dividing the maximum distance between superpoint $\hat{p}_i$ and all other superpoints $\hat{p}_j$ in the $\mathbf{\hat{P}}$ into $S$ segments. For a given layer $L=k$, the mask is defined as:

\begin{equation}
\mathbf M_{i, j}^{(L=k)} =
\begin{cases}
1, & \text{if } d_{i,j} \le \frac{k \cdot d_{i,\max}}{S}, \, k = 1, 2, \dots, S \\
0, & \text{otherwise}
\end{cases}
,
\end{equation}
where $ d_{i, \max} = \max_{\substack{i, j \in \mathbf{\hat P}}} d_{i,j}$, $k$ represents the segment index, with  $k = 1$  corresponding to the smallest distance range and $k = S$ corresponding to the largest. This enables the model to gradually expand the receptive field, focus on more refined local space in the early stage, reduce distant noise, and then learn multi-scale spatial feature expressions. After performing $S$ iterations of attention, the mixed superpoint features $\mathbf H^{\hat{P}}$ and $\mathbf H^{\hat{Q}}$ are obtained. 

\noindent\textbf{Superpoint Matching.}
To obtain the superpoint correspondences, we follow \cite{qin2022geometric}. First, the features $\mathbf H^{\hat{P}}$ and $\mathbf H^{\hat{Q}}$ obtained in the coarse stage are normalized to the unit sphere to yield the features $\mathbf{\bar{H}}^{\hat{P}}$ and $\mathbf{\bar{H}}^{\hat{Q}}$. Next, we compute a Gaussian correlation matrix $\mathbf{\hat{S}}$ with $\hat{s}_{i, j}=\exp \left(-\left\|\mathbf{\bar{h}}_{i}^{\hat{P}}-\mathbf{\bar{h}}_{j}^{\hat{Q}}\right\|_{2}^{2}\right)$. Following this, we perform dual normalization on $\mathbf{\hat{S}}$ to obtain $\mathbf{\bar{S}}$. Finally, based on the normalized results, the top-$k$ point pairs with the highest correlation scores are selected as superpoint correspondences $\hat{\mathcal{C}}=\left\{\left(\hat{p}_{i}, \hat{q}_{j}\right) \mid\left(\hat{p}_i, \hat{q}_j\right) \in \operatorname{topk}\left(\mathbf{\bar{S}}\right)\right\}$. 

\begin{figure*}[t]
    \centering
    \includegraphics [width=2.1\columnwidth]{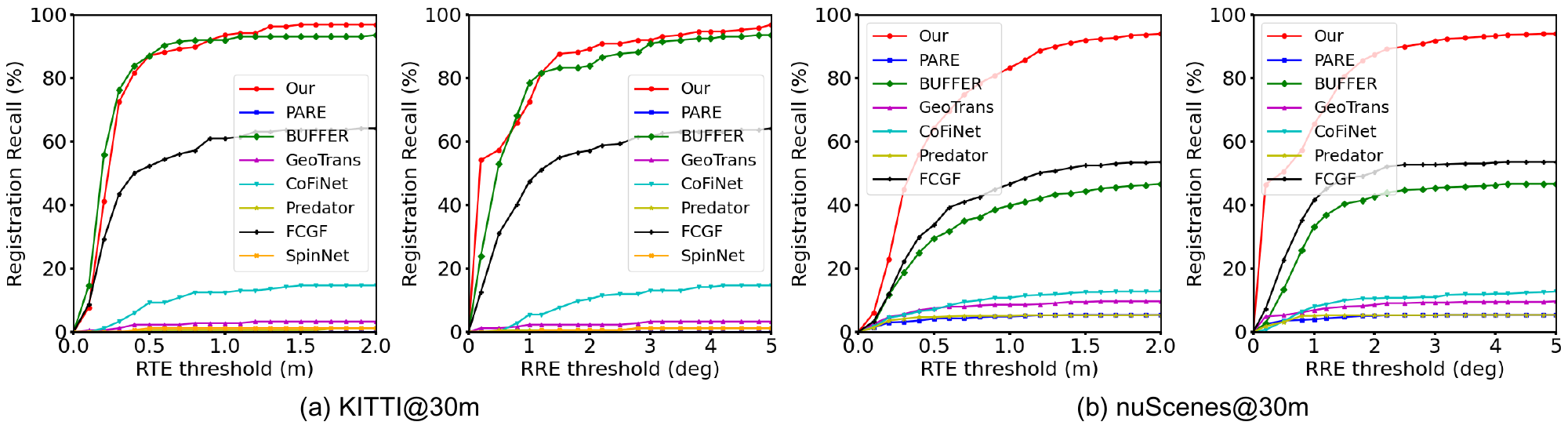}
    \vspace{-0.5 cm}
    \caption{Cross-distance registration recall results of different RRE and RTE thresholds on the KITTI@30m and nuScenes@30m.}
    \label{fig5}
\end{figure*}

\section{Experiments}
\label{sec:formatting}
\begin{table*} \centering
    \resizebox{\linewidth}{!}{
    \begin{tabular}{l|ccccc|ccccc|ccccc|ccccc|l}
        \toprule
        \textbf{Model} & \multicolumn{5}{c}{\textbf{KITTI@10m}} & \multicolumn{5}{c}{\textbf{KITTI@20m}} & \multicolumn{5}{c}{\textbf{KITTI@30m}} & \multicolumn{5}{c}{\textbf{KITTI@40m}} &  \\
        \cmidrule(lr){2-6} \cmidrule(lr){7-11} \cmidrule(lr){12-16} \cmidrule(lr){17-21}
        & RRE & RTE & RRE$^{*}$ & RTE$^{*}$ & RR & RRE & RTE & RRE$^{*}$ & RTE$^{*}$ & RR & RRE & RTE & RRE$^{*}$ & RTE$^{*}$ & RR & RRE & RTE & RRE$^{*}$ & RTE$^{*}$ & RR & mRR \\
        \midrule
        FCGF \cite{choy2019fully} & \textbf{0.196} & \textbf{0.060} & 1.455 & 0.248 & \underline{98.0} & \textbf{0.345} & \textbf{0.105} & 5.739 & 1.458 & 92.2 & \textbf{0.701} & \textbf{0.219} & 10.413 & 5.949 & 79.5 & 1.398 & \textbf{0.350} & \textbf{17.755} & 20.333 & 46.0 & 78.9\\
        Predator \cite{huang2021predator}& 0.415 & 0.093 & 1.086 & 0.254 & 97.8 & 1.267 & 0.249 & 9.003 & 7.430 & 59.4 & 1.881 & 0.279 & 19.460 & 28.709 & 0.5 & -- & -- & 26.632 & 38.956 & 0.0 & 39.4 \\
        CoFiNet \cite{yu2021cofinet}& 0.345 & 0.084 & 0.345 & 0.090 & \textbf{99.8} & 0.939 & 0.201 & 3.355 & 1.541 & 90.0 & 1.843 & 0.610 & 13.620 & 17.622 & 31.9 & 2.181 & 1.183 & 26.385 & 36.950 & 4.3 & 56.5 \\
        GeoTrans \cite{qin2022geometric}& 0.550 & 0.130 & 8.157 & 6.938 & 51.4 & 0.894 & 0.315 & 16.391 & 8.515 & 45.2 & 1.593 & 0.557 & 44.138 & 21.729 & 13.5 & 1.974 & 0.847 & 43.806 & 32.264 & 10.8 & 30.2 \\
        BUFFER \cite{ao2023buffer}& \underline{0.261} & \underline{0.065} & \textbf{0.264} & \textbf{0.072} & \textbf{99.8} & 0.508 & \underline{0.135} & \textbf{0.539} & \underline{0.349} & \underline{98.6} & 0.780 & \underline{0.265} & \textbf{4.873} & \underline{3.542} & \underline{87.0} & \underline{1.058} & \underline{0.396} & 21.435 & \underline{17.723} & \underline{53.2} & \underline{84.7} \\
        PARE \cite{yao2024pare}& 0.557 & 0.160 & 16.274 & 4.878 & 57.3 & 1.739 & 0.700 & 45.434 & 18.671 & 2.1 & -- & -- & 54.498 & 29.199 & 0.0 & -- & -- & 65.780 & 39.359 & 0.0 & 14.9\\
        \cdashline{0-21}
        GCL$^\dagger$ \cite{liu2023density}& 0.265 & 0.078 & 0.680 & 0.129 & 99.5 & 0.448 & 0.116 & 5.169 & 1.311 & 93.2 & 0.742 & 0.173 & 9.694 & 4.626 & 83.8 & 1.262 & 0.309 & 19.348 & 13.943 & 63.3 & 85.0\\
        \cdashline{0-21}
        UGP (Ours) & 0.282 & 0.078 & \underline{0.287} & \underline{0.085} & \textbf{99.8} & \underline{0.497} & 0.166 & \underline{1.098} & \textbf{0.309} & \textbf{98.9} & \underline{0.769} & 0.333 & \underline{5.287} & \textbf{1.867} & \textbf{90.8} & \textbf{0.975} & 0.523 & \underline{18.260} & \textbf{8.791} & \textbf{74.1} & \textbf{90.9}\\
        \bottomrule
    \end{tabular}
    }
    \caption{\textbf{Cross-dataset generalization experiments.} We \textbf{train on the nuScenes@10m} dataset and then test on the KITTI@10m, KITTI@20m, KITTI@30m and KITTI@40m. `$\dagger$' denotes that the method was trained using data within the range of [5m, 60m].}
    \label{tab:cross-dataset-distance}
\end{table*}
\subsection{Experimental Settings}
\noindent\textbf{Datasets.} We evaluated our method on two outdoor LiDAR datasets: KITTI (64-line LiDAR) \cite{kitti} and nuScenes (32-line LiDAR) \cite{nuscenes}. Following previous works \cite{choy2019fully, huang2021predator, yu2021cofinet}, sequences 0-5 are used for training, 6-7 for validation, and 8-10 for testing, with pairs created from scans at least 10m apart. To evaluate cross-distance and cross-dataset generalization, we expanded the test set to include pairs separated by 20m, 30m, and 40m. The goal is to train on the KITTI@10m dataset and test generalization on KITTI@20m, KITTI@30m and KITTI@40m. For nuScenes, we follow the common protocol \cite{nuscenes}, splitting it into training, validation, and testing subsets. Using the same approach, we train and validate on nuScenes@10m and test on nuScenes@20m, nuScenes@30m, and nuScenes@40m. To verify cross-dataset generalization, we trained on nuScenes@10m and tested on KITTI@10m, KITTI@20m, KITTI@30m, and KITTI@40m. Additional experiments are in Appendix Sec.~\ref{sec:Additional Experiments}.

\noindent\textbf{Metrics.} We follow the previous work \cite{huang2021predator,qin2022geometric, chen2024dynamic} and use five evaluation metrics: Patch Inlier Ratio (PIR), Inlier Ratio (IR), Relative Rotation Error (RRE), Relative Translation Error (RTE), Registration Recall (RR), and mean Registration Recall (mRR). RR in all experiments is based on the criteria RRE $ < 5^{\circ}$ and RTE $< 2$m. The definitions of the metrics can be found in Appendix Sec.~\ref{sec:Evaluation Metrics}.
\subsection{Generalization Experiment}
\noindent\textbf{Cross-Distance Generalization.} We compare the proposed UGP with other state-of-the-art methods in KITTI and nuScenes.  Experimental results are shown in Tab. \ref{tab:cross-distance}. Our method surpasses all the above approaches in terms of RR on both the KITTI and nuScenes datasets, particularly outperforming BUFFER. Compared to BUFFER, our method improves KITTI mRR by 6.2\% and KITTI@40m RR by 20.8\%. Furthermore, on the much sparser nuScenes dataset, our method significantly outperforms all other methods. Our method achieves SOTA at nuScenes@10m and then at nuScenes@20m improves by 42.3\% compared to GeoTrans, 14.7\% compared to BUFFER and 15\% compared to FCGF. At 30m our method still maintains 93.9\%, and outperforms BUFFER by 47.3\%. On the challenging 40m our method improves 57.1\% compared to BUFFER.

We further analyze the influence of point cloud pair distance changes on different methods. For coarse-to-fine methods (CoFiNet, GeoTransformer and PARE) performance drops significantly as point cloud pair distance changes. Even when generalizing to the relatively simpler 20m data subset, GeoTransformer’s RR drops to 7.5\%, while PARE’s RR drops to 1.8\%, and their performance continues to degrade as the distance increases. In contrast, for the point-wise or patch-wise methods (FCGF and BUFFER) , which avoid inter-frame information exchange in their network design, retain their generalization capacity. When the cross attention is eliminated (CoFiNet.w/o C, GeoTrans.w/o C and PARE.w/o C in Tab. \ref{tab:cross-distance}) , the generalization capability is unlocked. This further validates that the elimination of the cross-attention module plays a key role in enhancing the generalization of LiDAR scenes.

To evaluate the overall performance of our method, we also included average error metrics RRE$^{*}$ and RTE$^{*}$ for all test point pairs. Our method achieves the highest accuracy in terms of average RRE and RTE across all test point pairs for the 20m, 30m, and 40m distance generalization experiments. We display the cross-distance generalization registration recall with different RRE and RTE thresholds on two datasets in Fig. \ref{fig5}. According to this result, while the performance of BUFFER in KITTI@30m is higher than our method at some thresholds, we show a very large improvement at all thresholds in the larger nuScenes dataset test.
\begin{table}[ht] \centering
    \resizebox{0.9\linewidth}{!}{
    \begin{tabular}{lccccc}
        \toprule
        \textbf{Model} & \multicolumn{5}{c}{\textbf{nuScenes@10m}} \\
        \cmidrule(lr){2-6}
        & RRE($^{\circ}$)& RTE(m) & RRE$^*$($^{\circ}$) & RTE$^*$(m) & RR($\%$) \\
        \midrule
        FCGF \cite{choy2019fully}& \underline{0.390} & \underline{0.195} & 0.889 & 0.711 & 93.3  \\
        SpinNet \cite{ao2021spinnet}& 1.941 & 0.390 & 10.714 & 7.562 & 7.1 \\
        Predator \cite{huang2021predator}& 1.034 & 0.265 & 5.146 & 1.708 & 80.2  \\
        CoFiNet \cite{yu2021cofinet}& 0.713 & 0.231 & 0.959 & 1.122 & 88.5   \\
        GeoTrans  \cite{qin2022geometric}& 0.920 & 0.288 & 12.313 & 6.500 & 44.3   \\
        Buffer \cite{ao2023buffer}& \textbf{0.367} & \textbf{0.183} & \textbf{0.435} & \underline{0.626} & \underline{94.3} \\
        PARE \cite{yao2024pare}& 1.371 & 0.402 & 7.901 & 6.489 & 52.5 \\
        UGP(Ours) & 0.422 & 0.199 & \underline{0.478} & \textbf{0.322} & \textbf{98.4}\\
        \bottomrule
    \end{tabular}
    }
    \caption{The results of the cross-dataset generalization experiments \textbf{from KITTI@10m to nuScenes@10m}.}
    \label{tab:cross-dataset-nuscenes}
\end{table}

\noindent\textbf{Cross-Datasets Generalization.}
To evaluate the cross-dataset generalization capability of the proposed UGP, we conducted the following experiments: 1) from nuScenes to KITTI@ (10m, 20m, 30m and 40m), 2) from KITTI to nuScenes. As shown in Tab. \ref{tab:cross-dataset-distance}, the proposed UGP method, at KITTI@10m, achieved 99.8\% registration recall. Also on KITTI@20m, KITTI@30m and KITTI@40m, the best results 98.9\% (+0.3\%), 90.8\% (+3.8\%) and 74.1\% (+20.9\%) were achieved compared to BUFFER. It can be found that our method has a significant advantage at longer distances. Finally, we compared the GCL \cite{liu2023density} method using the pre-training weights of GCL in nuScenes. It can be found that our method also has a significant improvement in cross-dataset generalization compared to GCL, even though GCL utilizes [5m, 60m] point cloud frame during training. As shown in Tab. \ref{tab:cross-dataset-nuscenes}, our method achieves the best results despite the differences in sensor beams (KITTI: 64-line, nuScenes: 32-line). Compared to the sub-optimal BUFFER, the RR of our method improves by 4.1\%.

\begin{table} \centering
    \resizebox{\linewidth}{!}{
    \begin{tabular}{cccc|ccccccc}
        \toprule
        &&& & \multicolumn{7}{c}{\textbf{KITTI@30m}}\\
        \cmidrule(lr){5-11}
        No. & EC & PSA & BF & PIR & IR & RRE & RTE & RRE* & RTE* & RR \\
        \midrule
        a & - & - & - & 1.3 & 1.5 & 1.712 & 0.531 & 37.897 & 33.874 & 6.5  \\
        b & \checkmark & - & - & 49.7 & 39.4 & \underline{0.657} & \textbf{0.280} & 2.155 & 0.899 & 95.1 \\
        c & \checkmark & \checkmark & -  & 54.5 & 38.5 & 0.658 & 0.306 & \underline{2.090} & \textbf{0.541} & \underline{96.2}  \\
        d & \checkmark & - & \checkmark & 56.1 & \textbf{41.5} & \textbf{0.619} & 0.294 & \textbf{2.032} & \underline{0.681} & 95.7  \\
        e & \checkmark & \checkmark & \checkmark & \textbf{57.6} & \underline{40.6} & 0.677 & \underline{0.290} & 2.127 & 0.805 & \textbf{96.8} \\
        \midrule
        &&&  &\multicolumn{7}{c}{\textbf{KITTI@40m}} \\
        \cmidrule(lr){5-11}
        No. & EC & PSA & BF & PIR & IR & RRE & RTE & RRE* & RTE* & RR  \\
        \midrule
        a & - & - & - & 0.8 & 0.7 & 2.101 & 0.967 & 54.430 & 46.728 & 2.2 \\
        b & \checkmark & - & - & 24.2 & 18.2 & 0.950 & 0.486 & 21.539 & 10.401 & 66.2 \\
        c & \checkmark & \checkmark & -  & 28.3 & 18.5 & \underline{0.909} & \underline{0.470} & 17.692 & 9.091 & 71.2\\
        d & \checkmark & - & \checkmark  & \underline{29.9} & \textbf{20.3} &\textbf{0.907} & \textbf{0.461} & \underline{14.650} & \underline{8.967} & \underline{74.8} \\
        e & \checkmark & \checkmark & \checkmark  & \textbf{32.5} & \underline{19.9} & 0.975 & 0.494 & \textbf{13.095} & \textbf{6.168} & \textbf{82.0} \\
        \bottomrule
    \end{tabular}
    
    }
    \caption{\textbf{Ablation experiment} analysis of UGP on KITTI. Eliminating the cross-attention module (EC), progressive self-attention module (PSA), BEV fusion (BF) are ablated.}
    \label{tab:ablation-backbone}
\end{table}
\begin{figure}
    \centering
    \includegraphics [width=1\columnwidth]{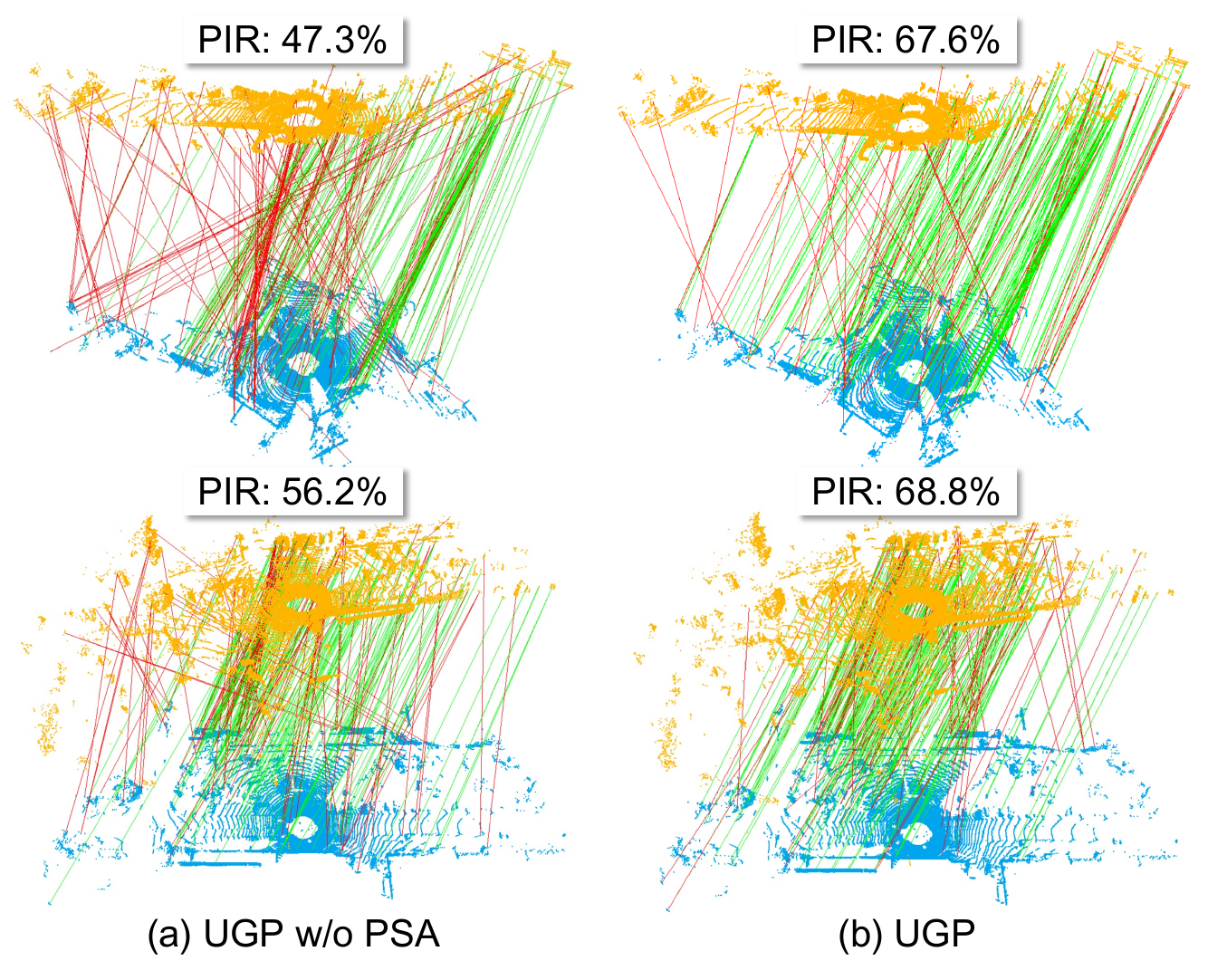}
    \caption{Registration results of the models with (a) vanilla self-attention and (b) progressive self-attention.}
    \label{fig:ab}
\end{figure}
\subsection{Ablation Study}
\noindent\textbf{Ablation of Backbone.}
First, we performed an ablation study on the core components of the method to demonstrate the effectiveness of each part, as shown in Tab. \ref{tab:ablation-backbone}. By comparing rows \textit{a-b} in Tab. \ref{tab:ablation-backbone}, we can see that EC significantly improves the network’s generalization capability. By comparing rows \textit{b-c}, we observe that PSA improves PIR by 4.8\% and increases RR on KITTI@30m by 1.1\% and by 5\% at longer distances (40m) compared to the vanilla self-attention module. we further visulize the matching results of the ablation PSA, as shown in Fig.~\ref{fig:ab}. It can be seen that the number of long-distance false matches is reduced significantly in the figure, proving that our PSA can effectively solve the feature ambiguity problem in vanilla self-attention. By comparing rows \textit{b-d} and \textit{c-e}, it can be observed that BEV provides semantic information about scene elements, reducing scene ambiguity. As a result, the framework’s PIR and RR both improve. Column \textit{e} in the table represents our complete proposed method, which achieves the best performance in the key metric of RR.

\begin{table}[ht] \centering
    \scalebox{0.59}{
        \begin{tabular}{l|c|c|c|c|c}
        \hline
        Estimators  & mRR & KITTI@10m & KITTI@20m & KITTI@30m & KITTI@40m \\
        \hline
        UGP+RANSAC50K & 92.4 & \textbf{99.8} & \textbf{99.3} & 95.1 & 75.5  \\
        UGP+MAC & \textbf{94.9} & \textbf{99.8} & \textbf{99.3} & \underline{96.2} & \textbf{84.2} \\
        UGP+LGR & \underline{94.5} & \textbf{99.8} & \textbf{99.3} & \textbf{96.8} & \underline{82.0} \\
        \hline
        \end{tabular}
    }
    \caption{\textbf{Comparison of cross-distance generalization performance} for different estimators on KITTI.}
    \label{tab:compare-estimator}
\end{table}

\noindent\textbf{Ablation of Estimator.}
We evaluated the impact of different estimators on the registration results, as shown in Tab. \ref{tab:compare-estimator}. The mRR of our method with RANSAC~\cite{fischler1981random} is 92.4\%, with MAC \cite{mac} is 94.9\%, and with LGR \cite{qin2022geometric} is 94.5\%.

\section{Conclusion}
\label{sec:conclusion}
In this paper, we revealed that inconsistent geometric representations in LiDAR scenarios cause the cross-attention module to limit network generalization capability. Based on this insight, we proposed UGP, a pruned framework designed to enhance generalization power for LiDAR point cloud registration. UGP eliminated cross-attention, introduced a progressive self-attention module and a BEV feature extraction module, enabling the network to prioritize local spatial associations and captured semantic information of scene elements. This reduced ambiguity in point clouds and boosted generalization performance. Extensive experiments showed that our method effectively addresses challenges across varying data distributions, including cross-distance and cross-dataset scenarios.

\noindent \textbf{Acknowledgments.} This work is supported in part by the National Natural Science Foundation of China (No.U23B2013 and 62372377).

{
    \small
    \bibliographystyle{ieeenat_fullname}
    \bibliography{main}
}
\clearpage
\setcounter{page}{1}
\maketitlesupplementary
\section{Method Details}
\label{sec:Method Details}
\noindent\textbf{Backbone.} Our Point-Encoder and Point-Decoder follow the KPConv-FPN~\cite{thomas2019kpconv} structure to perform point-level feature extraction. Before inputting the points into the Point-Encoder, we voxel downsample the KITTI point cloud with a voxel size of 0.3m and the nuScenes point cloud with a voxel size of 0.2m. The Point-Encoder and Point-Decoder follow GeoTrans~\cite{qin2022geometric}, with 5 and 3 layers, respectively. For the BEV-Encoder, we adopt a ResNet-like~\cite{resnet} structure with 3 layers.

\noindent\textbf{BEV Patch and Superpoint Indexing.} Given the superpoints, their 3D coordinates are projected onto the resolution $(H \times W)$ of the original BEV image. By accounting for the number of 2D max pooling operations $\beta$, we determine the index of each superpoint relative to the downsampled BEV patch. This correspondence establishes a one-to-one mapping between the features of the superpoints and BEV patches. Let $(u_i, v_i)$ represent the 2D coordinates of a superpoint in the BEV image. The corresponding index in the downsampled BEV feature map is calculated as:
\begin{equation}
u{\prime}_i = \left\lfloor \frac{u_i}{2^\beta} \right\rfloor, \quad v{\prime}_i = \left\lfloor \frac{v_i}{2^\beta} \right\rfloor.
\end{equation}
Here, $(u'_i, v'_i)$ represents the index of the patch feature in the downsampled BEV image. 


\noindent\textbf{Point Matching.} After obtaining the superpoint correspondences $\hat{\mathcal{C}}$, we follow a point-to-node assignment strategy~\cite{qin2022geometric} to uniquely assign dense points $\mathbf{\tilde{P}}$ and $\mathbf{\tilde{Q}}$ to their nearest superpoints, resulting in groups  $\mathcal{G}^P$  and  $\mathcal{G}^Q$ . Then, based on the superpoint correspondences $\hat{\mathcal{C}}$, we perform local dense matching. Unlike previous work, we input the attention features $\mathbf{\bar{H}}^{\hat{P}}$ and $\mathbf{\bar{H}}^{\hat{Q}}$ into the Point-Decoder to obtain dense point features $\mathbf{F}_{\tilde{P}} \in \mathbb{R}^{\tilde{n} \times \tilde{d}}$ and $\mathbf{F}_{\tilde{Q}} \in \mathbb{R}^{\tilde{m} \times \tilde{d}}$. For a given superpoint correspondence $\hat{\mathcal{C}}_{i}$  and its corresponding local dense points  $\mathcal{G}_{i}^{P} \text { and } \mathcal{G}_{i}^{Q}$ , we compute the cost matrix $\mathbf{\tilde{C}}$, where $\mathbf{\tilde{C}}_{i} = \mathbf{F}_{\tilde{P}}^{i}\left(\mathbf{F}_{\tilde{Q}}^{i}\right)^T / \sqrt{\tilde{d}}$. Then, we use the Sinkhorn algorithm \cite{sinkhorn1967concerning} to recompute the similarity matrix, resulting in  $\bar{\mathbf{C}}$. Based on $\bar{\mathbf{C}}$, we apply mutual top-$k$ to select the dense point correspondences. Finally, we gather all the dense point correspondences $\tilde{\mathcal{C}}_{i}$ for each coarse match in  $\hat{\mathcal{C}}$ , forming the final dense point correspondences $\mathcal{C} = \bigcup_{i=1}^{|\hat{\mathcal{C}}|} \tilde{\mathcal{C}}_i$. 

\noindent\textbf{Loss Function.}
Our framework's loss function consists of two components, $\mathcal{L}= \mathcal L_{c} + \mathcal L_{f}$, where $L_{c}$ and $L_{f}$ represent the same superpoint matching loss and point matching loss as \cite{qin2022geometric}.

\noindent\textbf{Implementation Details.} We conduct our experiments using PyTorch \cite{paszke2019pytorch} on an Intel (R) Xeon (R) Gold 5118 CPU and an NVIDIA RTX 3090 GPU. The Adam optimizer \cite{kingma2014adam} is used to train our model, with an initial learning rate of 1e-4 and a weight decay of 1e-6.

\begin{table*}[ht] \centering
    \resizebox{\linewidth}{!}{
    \begin{tabular}{l|ccccc|ccccc|ccccc|ccccc|l}
        \toprule
        \textbf{Model} & \multicolumn{5}{c}{\textbf{KITTI@10m}} & \multicolumn{5}{c}{\textbf{KITTI@20m}} & \multicolumn{5}{c}{\textbf{KITTI@30m}} & \multicolumn{5}{c}{\textbf{Train on KITTI@40m}} &  \\
        \cmidrule(lr){2-6} \cmidrule(lr){7-11} \cmidrule(lr){12-16} \cmidrule(lr){17-21}
        & RRE & RTE & RRE$^{*}$ & RTE$^{*}$ & RR & RRE & RTE & RRE$^{*}$ & RTE$^{*}$ & RR & RRE & RTE & RRE$^{*}$ & RTE$^{*}$ & RR & RRE & RTE & RRE$^{*}$ & RTE$^{*}$ & RR & mRR \\
        \midrule
        FCGF \cite{choy2019fully}& 1.180 & 0.285 & \underline{6.713} & 3.304 & 65.6 & 1.846 & 0.532 & 15.044 & 18.343 & 5.0 & -- & -- & 21.267 & 29.356 & 0.0 & 2.089 & \underline{0.524} & 23.732 & 39.583 & 0.7 & 17.8\\
        Predator \cite{huang2021predator}& 1.560 & 1.073 & 7.622 & 9.567 & 0.4 & -- & -- & \underline{14.587} & 19.495 & 0.0 & 1.873 & 0.607 & 18.341 & 29.323 & 0.5 & 2.037 & 1.165 & 22.941 & 38.708 & 0.7 & 0.4\\
        CoFiNet \cite{yu2021cofinet}& 1.325 & 0.280 & 7.400 & 3.420 & 61.4 & 1.899 & 0.588 & 23.895 & 17.138 & 8.5 & 1.897 & 0.678 & 28.712 & 24.868 & 10.3 & 2.375 & 0.888 & 23.776 & 25.417 & 23.7 & 26.0 \\
        GeoTrans \cite{qin2022geometric}& 1.296 & 0.300 & 38.783 & 19.247 & 42.2 & 1.133 & 0.319 & 16.224 & \underline{5.267} & \underline{68.7} & \underline{1.183} & \underline{0.375} & \textbf{5.524} & \textbf{2.217} & \textbf{80.5} & \textbf{1.037} & \textbf{0.514} & \textbf{8.639} & \textbf{3.235} & \textbf{85.6} & \underline{69.3} \\ 
        BUFFER \cite{ao2023buffer}& \underline{1.119} & \underline{0.182} & 8.679 & \underline{1.893} & \underline{81.6} & 2.202 & 0.435 & 36.861 & 12.615 & 30.2 & 2.747 & 0.556 & 70.534 & 27.838 & 3.2 & -- & -- & 76.648 & 39.030 & 0.0 & 28.8\\
        PARE \cite{yao2024pare}& 4.022 & 1.383 & 62.691 & 43.978 & 0.4 & \underline{0.951} & \underline{0.293} & 63.597 & 23.863 & 0.4 & 3.275 & \textbf{0.329} & 66.382 & 18.167 & 1.1 & 1.695 & 0.685 & 46.054 & 16.752 & 25.2 & 27.1\\
        UGP (Ours) & \textbf{0.384} & \textbf{0.117} & \textbf{0.749} & \textbf{0.263} & \textbf{98.6} & \textbf{0.794} & \textbf{0.249} & \textbf{3.652} & \textbf{1.613} & \textbf{89.7} & \textbf{1.139} & 0.378 & \underline{13.235} & \underline{5.616} & \underline{78.9} & \underline{1.385} & 0.596 & \underline{17.767} & \underline{8.927} & \underline{71.9} & \textbf{84.8}\\
        \bottomrule
    \end{tabular}
    }
    \caption{\textbf{Cross-distance generalization experiments.} \textbf{We train at KITTI@40m} and then test at 40m and nearer distances at 10m, 20m, and 30m. RRE and RTE denote the error for successfully matched point cloud pairs, while RRE$^*$ and RTE$^*$ reflect the error for all point cloud pairs, providing a more comprehensive evaluation. The final column shows the mean Registration Recall.
    }
    \label{tab:cross-distance-40}
\end{table*}

\begin{table*} \centering
    \resizebox{\linewidth}{!}{
    \begin{tabular}{l|ccccc|ccccc|ccccc|ccccc|l}
        \toprule
        \textbf{Model} & \multicolumn{5}{c}{\textbf{Train on KITTI@10m}} & \multicolumn{5}{c}{\textbf{KITTI@20m}} & \multicolumn{5}{c}{\textbf{KITTI@30m}} & \multicolumn{5}{c}{\textbf{KITTI@40m}} &  \\
        \cmidrule(lr){2-6} \cmidrule(lr){7-11} \cmidrule(lr){12-16} \cmidrule(lr){17-21}
        & RRE & RTE & RRE$^{*}$ & RTE$^{*}$ & RR & RRE & RTE & RRE$^{*}$ & RTE$^{*}$ & RR & RRE & RTE & RRE$^{*}$ & RTE$^{*}$ & RR & RRE & RTE & RRE$^{*}$ & RTE$^{*}$ & RR & mRR \\
        \midrule
        CoFiNet \cite{yu2021cofinet}& 0.699 & 0.175 & 1.912 & 0.731 & 94.2 & 1.739 & 0.488 & 9.901 & 8.917 & 46.6 & 1.934 & 0.878 & \underline{22.910} & 28.894 & 1.1 & - & - & \underline{24.768} & 39.194 & 0.0 & 35.5 \\
        GeoTrans \cite{qin2022geometric}&\textbf{0.291} & \textbf{0.082} & 0.358 & \textbf{0.095} & 99.3 & 2.453 & 0.815 & 36.227 & 26.501 & 2.1 & 4.298 & 1.232 & 44.197 & 35.805 & 0.5 & 2.229 & 0.855 & 48.724 & 45.975 & 1.4 & 25.8\\ 
        BUFFER \cite{ao2023buffer}& 0.309 & \underline{0.091} & \textbf{0.311} & \underline{0.097} & \textbf{99.8} & \underline{0.645} & \underline{0.188} & \underline{2.988} & \underline{1.554} & \underline{92.5} & \underline{0.997} & \textbf{0.291} & 24.145 & \underline{15.086} & \underline{51.4} & \underline{1.511} & \textbf{0.445} & 45.820 & \underline{30.337} & \underline{20.1} & \underline{66.0}\\
        UGP (Ours) & \underline{0.296} & 0.093 & \underline{0.336} & 0.110 & \underline{99.5} & \textbf{0.488} & \textbf{0.170} & \textbf{0.615} & \textbf{0.274} & \textbf{97.5} & \textbf{0.816} & \underline{0.354} & \textbf{5.348} & \textbf{1.879} & \textbf{90.3} & \textbf{1.091} & \underline{0.527} & \textbf{19.566} & \textbf{11.560} & \textbf{66.9} & \textbf{88.6} \\
        \bottomrule
    \end{tabular}
    }
    \caption{\textbf{Cross-distance generalization experiments on KITTI-Sparse.} \textbf{We train at KITTI@10m} and then \textbf{test at KITTI-Sparse@10m}
    and farther distances at \textbf{KITTI-Sparse@20m, KITTI-Sparse@30m, and KITTI-Sparse@40m}. KITTI-Sparse denotes that we use farthest point sampling (FPS) to downsample the input point clouds to \textbf{5000} points. RRE and RTE denote the error for successfully matched point cloud pairs, while RRE$^*$ and RTE$^*$ reflect the error for all point cloud pairs, providing a more comprehensive evaluation. The final column shows the mean Registration Recall.
    }
    \label{tab:cross-distance-sparse}
\end{table*}

\section{Evaluation Metrics}
\label{sec:Evaluation Metrics}
We follow previous work~\cite{qin2022geometric} and report Patch Inlier Ratio (PIR), Inlier Ratio (IR), Relative Rotation Error (RRE), Relative Translation Error (RTE) and Registration Recall (RR).

\noindent\textbf{Patch Inlier Ratio (PIR)} represents the proportion of superpoint (patch) matches that correctly overlap when aligned using the ground-truth transformation $\mathbf{T}_{\mathbf{P} \rightarrow \mathbf{Q}}$. This metric indicates the reliability and accuracy of the proposed superpoint (patch) correspondences:
\begin{equation}
\footnotesize
    \operatorname{PIR}=\frac{1}{|\hat{\mathcal{C}}|} \sum_{\left(\hat{p}_{i}, \hat{q}_{j}\right) \in \hat{\mathcal{C}}} \mathds{1}(\exists \tilde{\mathbf{p}} \in \mathcal{G}_{i}^{P}, \tilde{\mathbf{q}} \in \mathcal{G}_{i}^{Q} \text { s.t. }\|\mathbf{T}_{\mathbf{P} \rightarrow \mathbf{Q}}\left(\mathbf{\tilde{\mathbf{p}}}\right)-\tilde{\mathbf{q}}\|_{2}<\tau),  
\end{equation}
where $\tau=0.6$m and $\mathds{1}$ is the indicator function.

\noindent\textbf{Inlier Ratio (IR)} represents the proportion of inlier matches among all candidate point correspondences. A match qualifies as an inlier if the distance between the two points transformed by the ground-truth transformation  $\mathbf{T}_{\mathbf{P} \rightarrow \mathbf{Q}}$  is less than a threshold $\tau_1$ = 1.0m:
\begin{equation}
    \operatorname{IR}=\frac{1}{|\mathcal{C}|} \sum_{\left(\tilde{p}_{i}, \tilde{q}_{j}\right) \in \mathcal{C}} \mathds{1}\left(\left\|\mathbf{T}_{\mathbf{P} \rightarrow \mathbf{Q}}\left(\tilde{\mathbf{p}}_{i}\right)-\tilde{\mathbf{q}}_{i}\right\|_{2}<\tau_{1}\right).
\end{equation}

\noindent\textbf{Relative Rotation Error (RRE)} represents the geodesic distance measured in degrees between the estimated $\mathbf{R}_{est}$ and ground-truth $\mathbf{R}_{gt}$ rotation matrices. It quantifies the discrepancy between the predicted and actual rotation matrices:
\begin{equation}
\mathrm{RRE}=\arccos \left(\frac{\operatorname{trace}\left(\mathbf{R}^T_{est} \cdot \mathbf{R}_{gt}-1\right)}{2}\right).
\end{equation}

\noindent\textbf{Relative Translation Error (RTE)} represents the Euclidean distance between the estimated $\mathbf{t}_{est}$ and ground-truth $\mathbf{t}_{gt}$ translation vectors. This metric assesses the difference between the estimated and ground-truth translation vectors:

\begin{equation}
\mathrm{RTE}=\|\mathbf{t}_{est}-\mathbf{t}_{gt}\|_2.
\end{equation}

\noindent\textbf{Registration Recall (RR)} for all outdoor datasets is defined as the proportion of point cloud pairs where both RRE and RTE fall below specified thresholds (RRE $ < 5^{\circ}$ and RTE $< 2$m):
\begin{equation}
\mathrm{RR}=\frac{1}{M} \sum_{i=1}^M \mathds{1}\left(\mathrm{RRE}_i<5^{\circ} \wedge \mathrm{RTE}_i<2 \mathrm{m}\right),
\end{equation}
where $M$ is the number of point cloud pairs to be aligned.

\section{Additional Experiments}
\label{sec:Additional Experiments}

\subsection{Cross-distance (train on KITTI@40m)}
To comprehensively evaluate the generalization across different distances, we train on long-distance data (KITTI@40m) and generalize to shorter distances, as shown in Tab.~\ref{tab:cross-distance-40}. The results reveal that methods such as FCGF~\cite{choy2019fully}, Predator~\cite{huang2021predator}, CoFiNet~\cite{yu2021cofinet}, BUFFER~\cite{ao2023buffer}, and PARE~\cite{yao2024pare} struggle to achieve direct convergence at long distances. In contrast, the GeoTrans~\cite{qin2022geometric} network demonstrates the ability to converge at long distances and deliver good performance. However, its performance drops significantly when applied to simpler scenarios, such as KITTI@10m and KITTI@20m. This suggests that GeoTrans heavily relies on the visible data distribution, further highlighting that its cross-attention mechanism fails to adapt to variations in consistency representation of the same structure across different distances and datasets. Consequently, it cannot learn robust and generalizable features for LiDAR scenes. In contrast, our method not only successfully converges on KITTI@40m, but also gradually improves its performance as the distance decreases, consistent with the expected difficulty of the registration task. Ultimately, our method UGP achieves an mRR of 84.8\%, which is 15.5\% significantly ahead of the suboptimal GeoTrans.

\begin{table}[ht] \centering
    \resizebox{\linewidth}{!}{
    \begin{tabular}{lccccc}
        \toprule
        \textbf{Model} & \multicolumn{5}{c}{\textbf{Waymo@10m}} \\
        \cmidrule(lr){2-6}
        & RRE($^{\circ}$)& RTE(m) & RRE$^*$($^{\circ}$) & RTE$^*$(m) & RR($\%$) \\
        \midrule
        FCGF \cite{choy2019fully}& \textbf{0.137} & 0.081 & 0.597 & 0.155 & \underline{99.2}  \\
        SpinNet \cite{ao2021spinnet}& 0.377 & 0.096 & 0.553 & 0.171 & \underline{99.2} \\
        Predator \cite{huang2021predator}& 0.190 & 0.082 & 0.190 & 0.082 & \textbf{100.0}  \\
        CoFiNet \cite{yu2021cofinet}& 0.179 & \underline{0.080} & 0.179 & \underline{0.080} & \textbf{100.0}   \\
        GeoTrans  \cite{qin2022geometric}& 0.255 & 0.124 & 3.740 & 7.247 & 61.5   \\
        Buffer \cite{ao2023buffer}& \underline{0.171} & 0.088 & \underline{0.171} & 0.088 & \textbf{100.0} \\
        PARE \cite{yao2024pare}& 0.270 & 0.136 & 1.051 & 4.003 & 76.9 \\
        UGP (Ours) & \textbf{0.137} & \textbf{0.075} &  \textbf{0.137} & \textbf{0.075} & \textbf{100.0}\\
        \bottomrule
    \end{tabular}
    }
    \caption{The results of the cross-dataset generalization experiments \textbf{from KITTI@10m to Waymo@10m}.}
    \label{tab:cross-dataset-waymo}
\end{table}

\subsection{Cross-dataset (KITTI@10m to Waymo@10m)}
To comprehensively evaluate the cross-dataset generalization ability of our method, we supplemented the results with training on KITTI@10m and testing on Waymo@10m. For the Waymo dataset (64-line LiDAR), we follow the same protocol in \cite{sun2020scalability} and utilize the testing subset, the results are shown in Tab.~\ref{tab:cross-dataset-waymo}. Our method achieved a state-of-the-art RR of 100\% and the lowest errors in both RRE and RTE.
\subsection{KITTI-Sparse}
To evaluate the robustness of the network to extremely sparse LiDAR point clouds, we use farthest point sampling (FPS) to downsample the input point clouds to \textbf{5000} points, which is referred to as KITTI-Sparse. We compare CoFiNet~\cite{yu2021cofinet}, GeoTrans~\cite{qin2022geometric}, and BUFFER~\cite{ao2023buffer}, with the results shown in Tab.~\ref{tab:cross-distance-sparse}. UGP demonstrates a significant advantage at 20m, 30m, and 40m. Compared to the suboptimal BUFFER, UGP achieves 97.5\% (+5.0\%) RR at 20m, 90.3\% (+38.9\%) RR at 30m, and 66.9\% (+46.8\%) RR at 40m.

\begin{table*}[ht] \centering
    \resizebox{\linewidth}{!}{
    \begin{tabular}{c|c|ccccc|ccccc|ccccc|ccccc|l}
        \toprule
        & & \multicolumn{5}{c}{\textbf{Train on KITTI@10m}} & \multicolumn{5}{c}{\textbf{KITTI@20m}} & \multicolumn{5}{c}{\textbf{KITTI@30m}} & \multicolumn{5}{c}{\textbf{KITTI@40m}} &  \\
        \cmidrule(lr){3-7} \cmidrule(lr){8-12} \cmidrule(lr){13-17} \cmidrule(lr){18-22}
        \textbf{Noise $\sigma$} & \textbf{Method} & RRE & RTE & RRE$^{*}$ & RTE$^{*}$ & RR & RRE & RTE & RRE$^{*}$ & RTE$^{*}$ & RR & RRE & RTE & RRE$^{*}$ & RTE$^{*}$ & RR & RRE & RTE & RRE$^{*}$ & RTE$^{*}$ & RR & mRR \\
        \midrule
        \multirow{3}{*}{\textbf{0.01}} 
        & FCGF~\cite{choy2019fully} & \textbf{0.214} & \textbf{0.061} & 0.868 & 0.153 & 98.9 & \textbf{0.389} & \textbf{0.128} & 5.507 & 1.323 & 92.8 & 1.080 & 0.415 & 11.739 & 10.998 & 62.0 & 1.634 & 0.835 & 20.555 & 30.727 & 18.7 & 68.1 \\
        & BUFFER~\cite{ao2023buffer} & 0.266 & 0.073 & \textbf{0.269} & 0.079 & \textbf{99.8} & 0.472 & \textbf{0.128} & 0.520 & 0.280 & 98.6 & 0.669 & \textbf{0.237} & \textbf{1.366} & 1.876 & 93.0 & 1.037 & \textbf{0.363} & 13.327 & 15.820 & 61.1 & 88.1 \\
        & UGP (ours) & 0.243 & 0.071 & 0.294 & \textbf{0.078} & \textbf{99.8} & 0.399 & 0.144 & \textbf{0.441} & \textbf{0.219} & \textbf{99.3} & \textbf{0.641} & 0.282 & 2.206 & \textbf{1.052} & \textbf{95.7} & \textbf{0.994} & 0.478 & \textbf{12.938} & \textbf{7.174} & \textbf{78.4} & \textbf{93.3}\\
        \midrule
        \multirow{3}{*}{\textbf{0.03}}
        & FCGF~\cite{choy2019fully} & \textbf{0.204} & \textbf{0.060} & 0.908 & 0.158 & 98.7 & \textbf{0.374} & \textbf{0.117} & 6.149 & 1.731 & 91.4 & 1.014 & 0.396 & 11.600 & 11.886 & 58.2 & 1.648 & 0.728 & 22.072 & 31.061 & 17.9 & 66.6 \\
        & BUFFER~\cite{ao2023buffer} & 0.277 & 0.074 & \textbf{0.278} & 0.080 & \textbf{99.8} & 0.478 & 0.127 & 0.508 & 0.342 & 98.6 & 0.720 & \textbf{0.243} & 3.293 & 2.026 & 93.0 & 0.959 & \textbf{0.358} & 18.489 & 17.286 & 55.4 & 86.7\\
        & UGP (ours) & 0.246 & 0.070 & 0.292 & \textbf{0.077} & \textbf{99.8} & 0.419 & 0.147 & \textbf{0.461} & \textbf{0.222} & \textbf{99.3} & \textbf{0.607} & 0.285 & \textbf{2.054} & \textbf{1.180} & \textbf{95.7} & \textbf{0.880} & 0.451 & \textbf{16.181} & \textbf{6.449} & \textbf{77.0} & \textbf{93.0}\\
        \midrule
        \multirow{3}{*}{\textbf{0.05}}
        & FCGF~\cite{choy2019fully} & \textbf{0.216} & \textbf{0.061} & 1.205 & 0.235 & 98.6 & \textbf{0.422} & 0.130 & 5.774 & 1.839 & 91.0 & 1.248 & 0.410 & 11.973 & 14.004 & 53.3 & 1.601 & 1.004 & 19.855 & 31.803 & 17.0 & 65.0 \\
        & BUFFER~\cite{ao2023buffer} & 0.287 & 0.073 & 0.289 & 0.080 & \textbf{99.8} & 0.496 & \textbf{0.129} & 0.544 & 0.280 & 98.6 & 0.790 & \textbf{0.252} & \textbf{3.491} & 2.286 & 91.9 & 1.119 & \textbf{0.397} & 18.584 & 17.225 & 51.8 & 85.5\\
        & UGP (ours) & 0.253 & 0.070 & \textbf{0.285} & \textbf{0.077} & \textbf{99.8} & 0.427 & 0.150 & \textbf{0.460} & \textbf{0.224} & \textbf{99.3} & \textbf{0.597} & 0.283 & 3.554 & \textbf{1.214} & \textbf{93.5} & \textbf{1.101} & 0.488 & \textbf{14.526} & \textbf{8.571} & \textbf{75.5} & \textbf{92.0}\\
        \bottomrule
    \end{tabular}
    }
    \caption{\textbf{Comparison of results under varying noise intensities}, with $\sigma$ representing the standard deviation. RRE and RTE denote the error for successfully matched point cloud pairs, while RRE$^*$ and RTE$^*$ reflect the error for all point cloud pairs, providing a more comprehensive evaluation. The final column shows the mean Registration Recall.}
    \label{tab:kitti-noise}
\end{table*}
\begin{figure}[ht]
    \centering
    \includegraphics [width=0.95\linewidth]{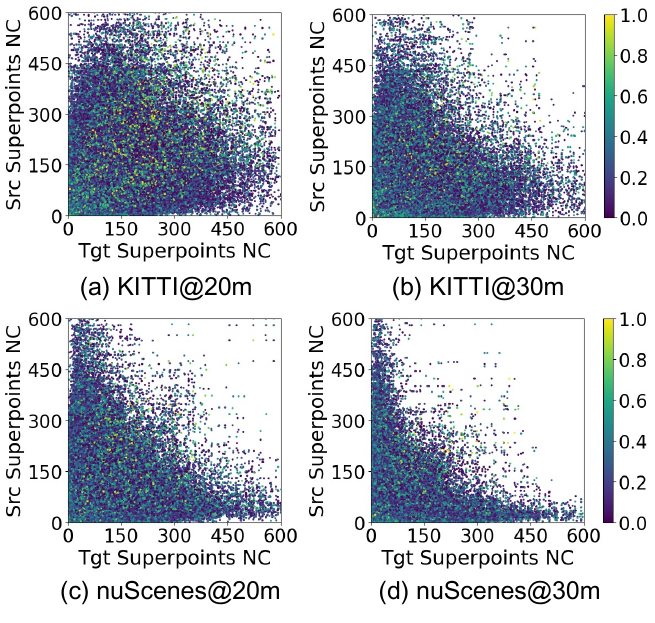}
    \caption{\textbf{Visualization of data distributions for ground truth matching point pairs across varying distances and datasets}. (a-d) Each point in the figure represents a ground truth corresponding superpoint pair. The position of each point indicates the neighborhood count (NC) of the superpoint within a radius of $r=2.4$m in both the source (src) and target (tgt) point clouds, and the color represents the overlap degree of the corresponding superpoint pairs after rotation by the ground truth transformation.}
    \label{fig:density2030}
\end{figure}
\subsection{KITTI-Noise}
To evaluate the robustness of our method to noise in real-world environments, we add Gaussian-distributed random noise $N\left(0,\sigma ^{2}\right)$ (clipped to $\left[ -3\sigma, +3\sigma \right]$) to each point’s position to simulate the measurement errors and noise encountered by LiDAR sensors in real-world scenarios, as shown in Tab.~\ref{tab:kitti-noise}. Since other methods almost completely fail at long ranges, we only compare our method with FCGF~\cite{choy2019fully} and BUFFER~\cite{ao2023buffer}. Our method achieves the highest RR across different levels of noise. Notably, the RR of our method remains unaffected at short distances, such as KITTI@10m and KITTI@20m. At KITTI@30m, the RR decreases by 3.3\% under a noise level of $\sigma = 0.05$. At KITTI@40m, the decrease reaches 6.5\% under the same noise level. In summary, our method experiences an mRR reduction of no more than 2.5\% (from 94.5\% to 92.0\%) at an intensity of $\sigma=0.05$, demonstrating a certain level of robustness to noise.

\begin{table}[ht]\centering
    \scalebox{0.75}{
        \begin{tabular}{c|c|c}
        \hline
        LayerNum  & KITTI@30m (RR\%) & KITTI@40m (RR\%)\\
        \hline
        2 & 95.1 & \underline{80.6}\\
        3 & \textbf{96.8} & \textbf{82.0}\\
        4 & 95.7 & 74.8\\
        5 & \underline{96.2} & 72.7\\
        6 & 95.7 & 72.7\\
        \hline
        \end{tabular}
        }
    \caption{\textbf{Ablation experiment} of progressive self-attention module partitioning \textbf{with different number of spatial layers $L$}. $L=3$ is selected to achieve the highest RR on KITTI@30m and KITTI@40m.} 
    \label{tab:ablation-parameter}
\end{table}

\subsection{Ablation of Parameter}
We conducted an ablation study on the number of layers $L$ used to divide the space in the PSA. As shown in Tab. \ref{tab:ablation-parameter}, we selected $L=3$, which provided the best performance, as the final network implementation.

\subsection{Mechanism Analysis}
Admittedly, cross-attention achieves promising performance under same-distance/dataset settings. However, for LiDAR registration requiring cross-domain generalization, we identify a fundamental limitation: Cross-attention learns static density matching patterns from training data but struggles to extrapolate to the real physical law of LiDAR density decay ($\rho \propto 1 / d^{2}$). When applied to cross-distance or cross-dataset scenarios, the learned correlation patterns become invalid due to density scaling or differences in LiDAR type. 

\begin{figure}[ht]
    \centering
    \includegraphics [width=1\columnwidth]{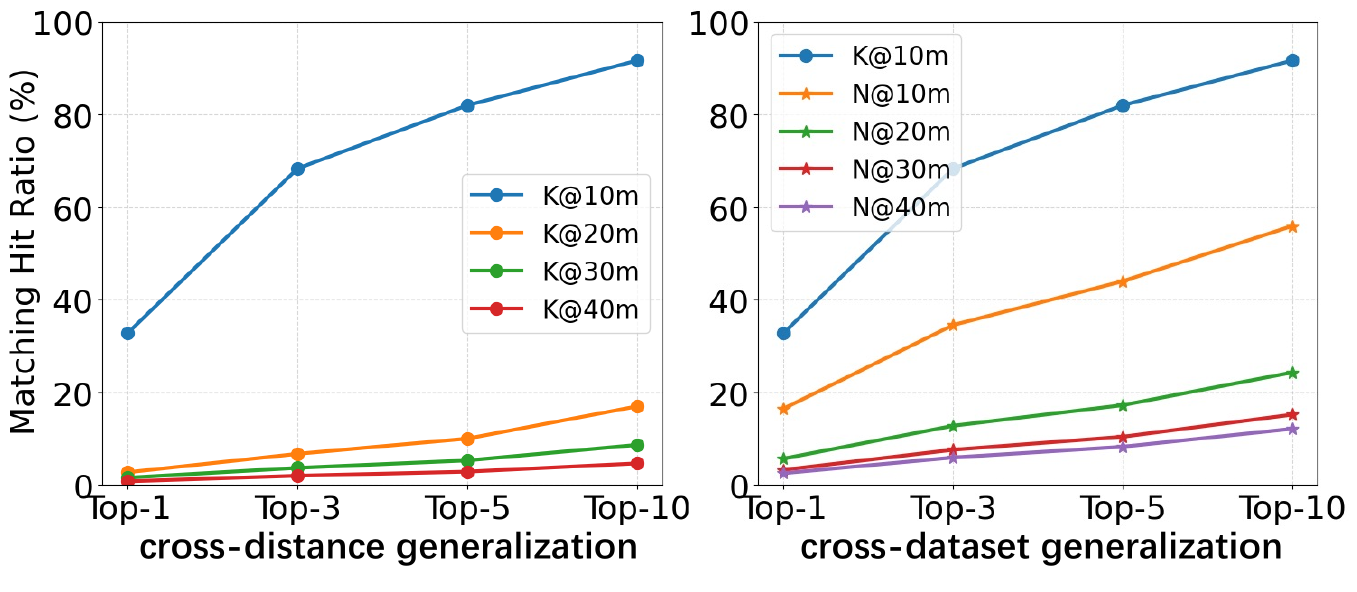}
    \caption{\textbf{Visualization of the matching hit ratio} in cross-distance and cross-dataset generalization experiments.}
    \label{fig:QK}
\end{figure}

To this end, we analyzed the $\operatorname{Softmax}(Q K^{T}/{\sqrt{d}})$ mechanism in UGP w. cross-attention, as shown in Fig.~\ref{fig:QK}. At K@10m training, regions with high cross-attention scores (Top-$10$) covered 91.64\% of true matches, \emph{validating the effectiveness of the $Q K^{T}$ mechanism in capturing point cloud correspondences and guiding feature updates}. However, cross-distance or cross-dataset scenarios exhibit LiDAR distribution shifts, causing true matches in high-score regions to plummet. This introduces false matches, increases feature ambiguity, and weakens generalization.

\vspace{-10pt}
\vspace{-16pt}

\section{Visualizations}
\label{sec:Visualizations}

\noindent\textbf{LiDAR Point Cloud Registration Characteristics.} To supplement Fig.~\ref{fig:motivation} (a) in Sec.~\ref{sec:Motivation}, we provide additional details. Specifically, Fig.~\ref{fig:density2030} illustrates the data distributions of ground truth matching point pairs for KITTI and nuScenes at distances of 20m and 30m.

\begin{figure*}
    \centering
    \includegraphics [width=1\linewidth]{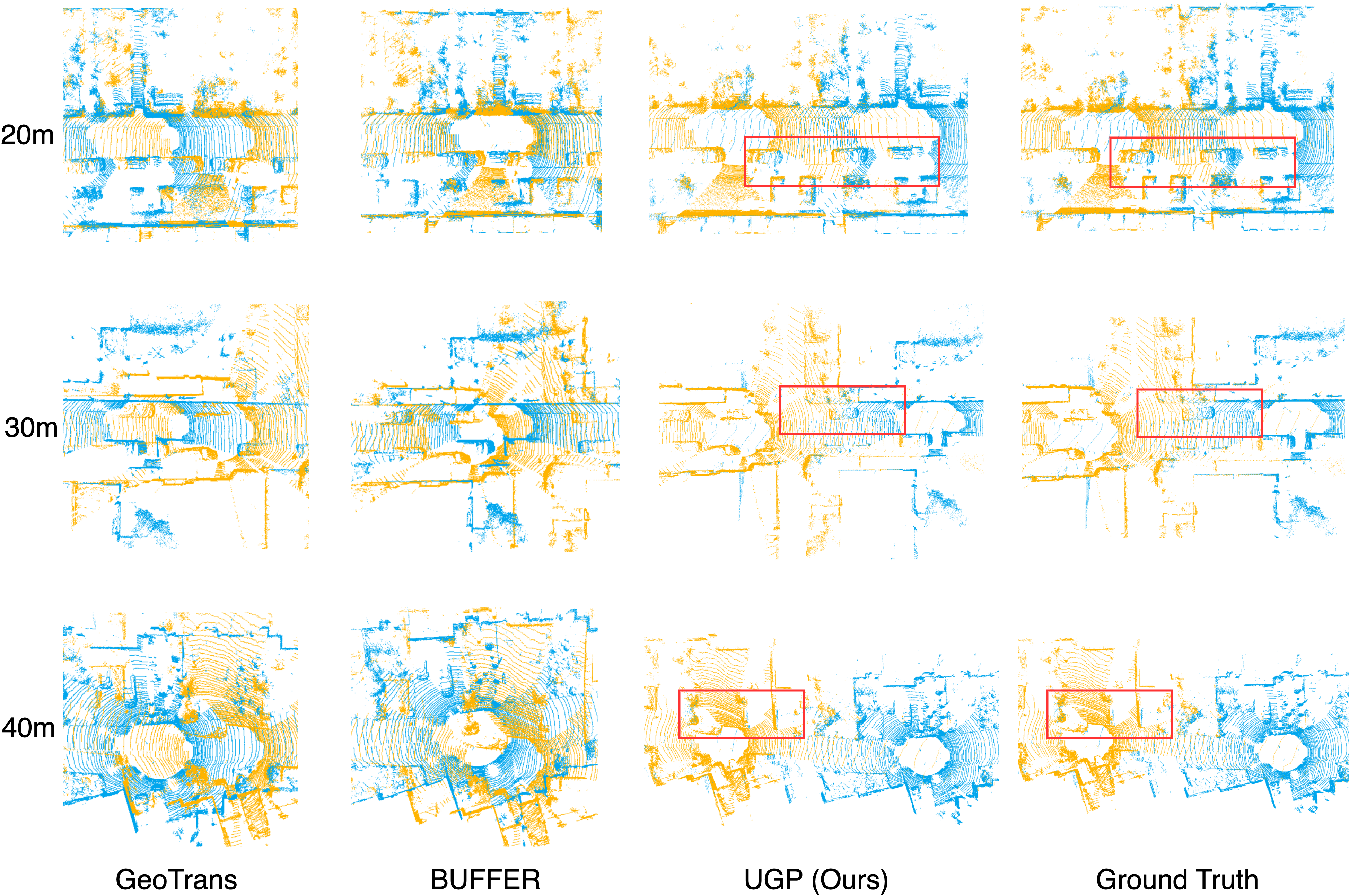}
    \caption{\textbf{Cross-distance generalization visualization} of GeoTrans~\cite{qin2022geometric}, BUFFER~\cite{ao2023buffer}, and UGP \textbf{on the KITTI~\cite{kitti}} dataset. Each row shows the point cloud pair matching results at different distances.}
    \label{fig:vis_cd_kitti}
\end{figure*}

\begin{figure*}
    \centering
    \includegraphics [width=1\linewidth]{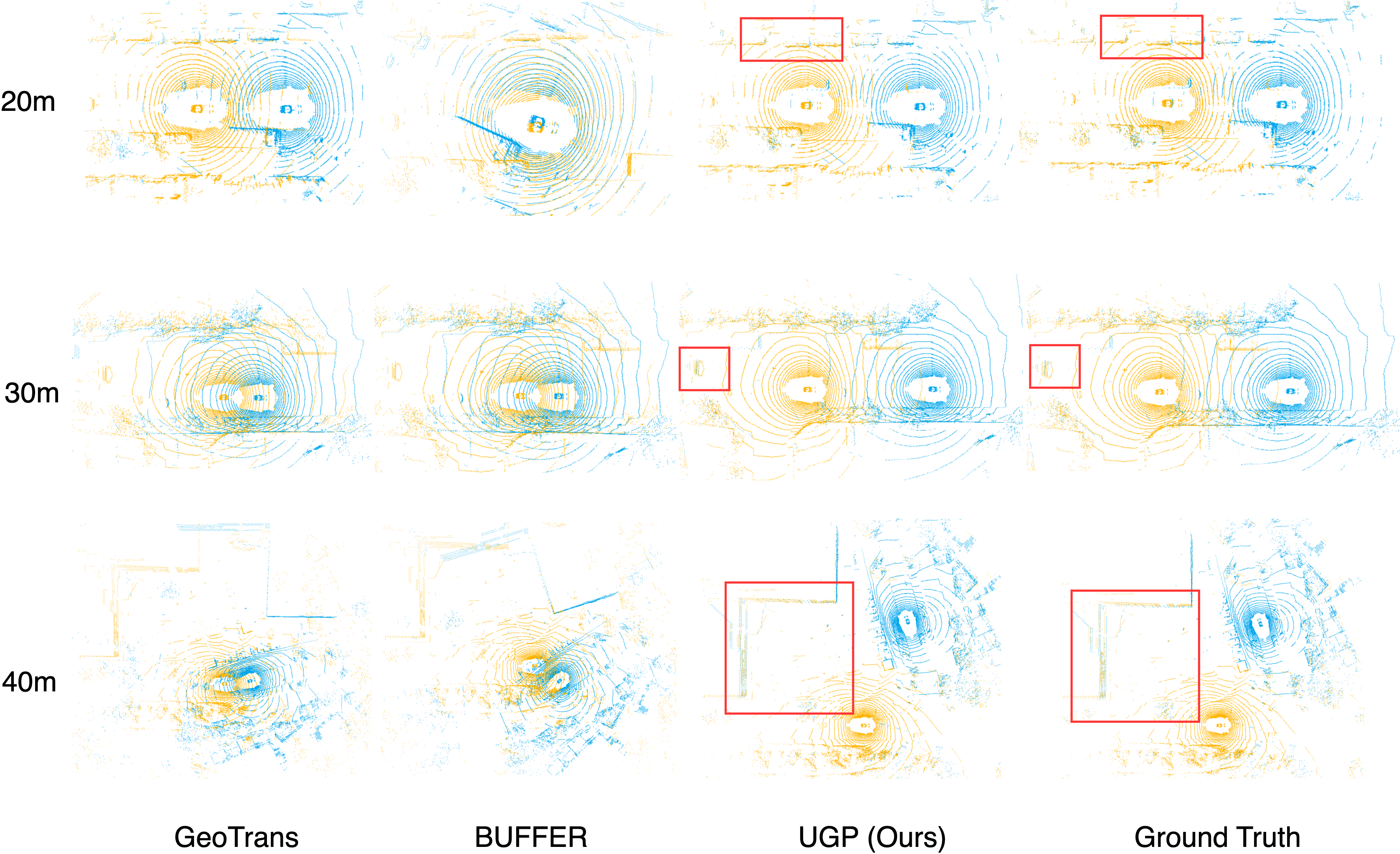}
    \caption{\textbf{Cross-distance generalization visualization} of GeoTrans~\cite{qin2022geometric}, BUFFER~\cite{ao2023buffer}, and UGP \textbf{on the nuScenes~\cite{nuscenes}} dataset. Each row presents the point cloud pair matching results at different distances.}
    \label{fig:vis_cd_nuscenes}
\end{figure*}

\noindent\textbf{Registration Results.} The cross-distance registration results for KITTI and nuScenes are shown in Fig.~\ref{fig:vis_cd_kitti} and Fig.~\ref{fig:vis_cd_nuscenes}.

\end{document}